%% file: main.tex
\documentclass{IEEEtaes}

\usepackage{color,array,amsthm}
\usepackage{graphicx}
\usepackage{amsmath}
\usepackage{amssymb}
\usepackage{booktabs}
\usepackage{bm}
\usepackage{placeins}
\usepackage{float}
\usepackage{url}
\usepackage{etoolbox}

\makeatletter
\patchcmd{\HyOrg@maketitle}
  {0018-9251\ \copyright\ \@pubyear\ IEEE}
  {{\scriptsize This work has been submitted to the IEEE for possible
  publication. Copyright may be transferred without notice, after which this
  version may no longer be accessible.}}
  {}
  {\PackageError{main}{Could not remove the IEEE copyright line}{}
  }

\def\ps@headings{%
  \def\@oddhead{}%
  \def\@evenhead{}%
  \def\@oddfoot{\hbox to \textwidth{{\rffont
    \leftmark\ifCLASSOPTIONcorrespondence\else
    \ifCLASSOPTIONletter\else:\fi\fi\ \rightmark}%
    \hfill{\rffont\thepage}}}%
  \def\@evenfoot{\hbox to \textwidth{{\rffont\thepage}\hfill
    {\rffont\rightmark}}}%
}
\def\ps@plain{%
  \def\@oddhead{}%
  \def\@evenhead{}%
  \def\@oddfoot{\hbox to \textwidth{\hfill{\rffont\thepage}}}%
  \let\@evenfoot\@oddfoot
}
\makeatother
\begin{document}

\title{VISTA: An Attention-Based Multi-Agent Reinforcement Learning Architecture for Space Situational Awareness Sensor Tasking}

\author{Miguel Leiva-V\'elez}
\affil{ETSIAE-School of Aeronautics, Universidad Politécnica de Madrid, Spain} 

\author{Adalberto Claudio Quiros}
\affil{Indra Sistemas S.A., Spain}

\author{Nicolas Gaston Rozado}
\affil{Indra Sistemas S.A., Spain}

\author{Hodei Urrutxua}
\affil{Escuela de Ingeniería de Fuenlabrada, Universidad Rey Juan Carlos, Spain}

\author{V\'ictor Rodr\'iguez-Fern\'andez}
\affil{Department of Computer Systems Engineering, Universidad Politécnica de Madrid, Spain}

\receiveddate{This work has been partially supported by the Spanish Agencia Estatal de Investigación (AEI) under Grants PID2024-161963OB-C22 and PID2024-161963OB-C21 (Coordinated Project ACTIVATION) funded by MICIU/ AEI / 10.13039/501100011033 / FEDER, UE. This work has also been supported by the Madrid Government (Comunidad de Madrid-Spain) under the Multiannual Agreement 2023-2026 with Universidad Polit\'ecnica de Madrid in the Line A, Emerging PhD researchers.}
\corresp{{\itshape Corresponding author: Miguel Leiva-V\'elez.}}
\authoraddress{Miguel Leiva-V\'elez is with ETSIAE-School of Aeronautics,
Universidad Polit\'ecnica de Madrid,  Pl. del Cardenal Cisneros, 3, 28040 Madrid, Spain (e-mail: \href{mailto:miguel.leiva.velez@alumnos.upm.es}%
{miguel.leiva.velez@alumnos.upm.es}). Adalberto Claudio Quiros and Nicolas
Gaston Rozado are with Indra Sistemas S.A., Alcobendas, Spain. Hodei Urrutxua is
with Escuela de Ingenier\'ia de Fuenlabrada, Universidad Rey Juan Carlos,
Fuenlabrada, Spain. V\'ictor Rodr\'iguez-Fern\'andez is with the Department of
Computer Systems Engineering, Universidad Polit\'ecnica de Madrid, Madrid,
Spain.}
\supplementary{Code and frozen experiment configurations are available at
\url{https://github.com/RocketNeurons/VISTA-SSA}.}

\markboth{LEIVA-V\'ELEZ ET AL.}{VISTA: Attention-Based MARL for SSA Sensor Tasking}
\maketitle

\begin{abstract}
\input{Sections/abstract}
\end{abstract}


\section{INTRODUCTION}
\input{Sections/introduction}

\section{PROBLEM FORMULATION}\label{sec:problem-formulation}
\input{Sections/problem_formulation}

\section{LEARNING FRAMEWORK}\label{sec:learning-paradigm}
\input{Sections/learning_framework}

\section{NEURAL ARCHITECTURE}
\input{Sections/neural_architecture}

\section{EXPERIMENTAL SETUP}\label{sec:setup}
\input{Sections/experimental_setup}

\section{RESULTS AND DISCUSSION}\label{sec:results}
\input{Sections/results_and_discussion}

\section{CONCLUSION}
\input{Sections/conclusion}

\section*{Acknowledgment}
The authors used Codex and Claude for language and grammar editing and for code
review and improvement. All AI-assisted changes were reviewed and validated by
the authors, who take full responsibility for the manuscript and software.

\FloatBarrier

\bibliographystyle{IEEEtaes}
\bibliography{references}

\input{Profiles/biography}
\end{document}

%% file: Sections/abstract.tex
The rapid growth of resident space objects is increasing the complexity of space situational awareness sensor tasking, challenging classical optimization methods as they allocate finite, heterogeneous, and distributed sensing resources across ever-larger catalogues. Existing deep reinforcement learning approaches show promise in reduced settings, but fixed-dimensional state and action representations limit their ability to scale to large, dynamic catalogues and distributed sensing networks. We introduce VISTA (Variable-Entity Intelligent Sensor Tasking Architecture), a scalable deep reinforcement learning architecture for persistent uncertainty-driven catalogue maintenance across variable object populations and sensor configurations. VISTA combines physics- and mission-informed top-$K$ retrieval with entity-centric attention, recurrent memory, and pointer-based action decoding, thereby keeping each agent's observation and action spaces independent of catalogue size. We evaluate VISTA across different scenarios, from fixed-size single-sensor benchmarks to large-scale space-based tasking and heterogeneous cooperative sensing. With 30 orbiting targets, VISTA recovers the catalogue 31.2\% faster than the fixed-dimensional recurrent baseline. In the large-scale regime, VISTA reduces five-hour uncertainty by 97.5\% relative to the strongest classical reference and by 99.3\% relative to the recurrent learner. Zero-shot tests up to 20,000 objects reveal near-linear relations between sensing capacity, catalogue size, and recovery horizon. Learned policies also exhibit sensor modality adaptation and generalization to population and initial-uncertainty shifts. Together, these results demonstrate that VISTA provides a scalable framework for adaptive space situational awareness sensor tasking across large, distributed networks of heterogeneous ground- and space-based sensors.

%% file: Sections/introduction.tex
The intensifying pace of space activity is driving a significant increase in the number, diversity and deployment rate
of resident space objects (RSOs). In this incrementally crowded environment, ensuring the long-term
sustainability of space operations will require space situational awareness (SSA) systems capable of coordinating large-scale, heterogeneous networks of ground- and space-based sensors, efficiently translating catalogue and service requirements into assignments for optical, radio-frequency, radar, and laser-ranging assets. This work focuses on that decision-layer, while abstracting measurement acquisition, detection and tracklet formation, data association and precise orbit determination.

Sensor tasking determines which RSOs should be observed by which sensors at any given time. As the number of sensors and targets grows, this problem becomes a highly complex, time-dependent combinatorial optimization task, constrained by orbital dynamics, sensor-specific limitations, geometric visibility, and mission-driven objectives. Policies driven only by immediate value may favour locally attractive targets while less observable or rapidly diverging objects become poorly maintained \cite{erwin2010dynamic,williams2013coupling,linares2017autonomous}. Reliable catalogue maintenance consequently requires balancing physical fidelity, planning horizon, adaptability, and computational cost \cite{vallado2011simulating,little2020space,siew2022space}. 

Classical SSA tasking has used covariance reduction \cite{miller2007new,hill2010covariance}, information gain \cite{kreucher2004information}, coordinated assignment \cite{erwin2010dynamic,williams2013coupling}, dynamic programming \cite{li2009approximate}, stochastic control \cite{gentile2023stochastic,CAI2020234}, and receding-horizon planning \cite{sunberg2015information}. Such methods can reach operational scale: Ravago and Jones close the loop for 4,545 LEO objects \cite{ravago2021risk}, while Shtofenmakher and Balakrishnan schedule approximately 20,000 resident space objects with 24 sensors by exploiting sparse feasibility \cite{shtofenmakher2025optimal}. Cislunar POMDPs \cite{chaturvedi2025information}, binary programs \cite{shimane2025multi}, and Beam A* \cite{federici2024optimal} provide further alternatives. These methods exploit prescribed structure: opportunities are precomputed, variables rebuilt for each scenario, or candidate values and search trees expensively reevaluated online. Their tractable formulations often rely on simplified scores or bounded searches that may miss subtle nonlinear dependencies and delayed consequences, while changes in catalogue, geometry, or objective may require new preprocessing or reformulation.

Deep reinforcement learning (DRL) offers a complementary approach by shifting policy optimization to offline training and reducing online decision-making to neural inference, rather than repeatedly solving or reformulating the tasking problem. Additionally, DRL combines neural networks' capacity to learn expressive representations of high-dimensional inputs with reinforcement learning's ability to optimize sequential, long-horizon objectives \cite{sutton_reinforcement_2018}. This shift, however, does not by itself resolve the representational and computational challenges of scaling. Existing SSA sensor tasking policies trained through deep reinforcement learning usually assume fixed sensor and target counts or fixed observation and action dimensions \cite{linares2016dynamic,linares2017autonomous,roberts2021deep,siew2022optimal,siew2022space,oakes2022double,mignocchi2026meta,huterer2025reinforcement}. Extending such representations to larger catalogues creates a fundamental tradeoff. Encoding the full catalogue makes network size and computation scale with population, although many objects are inaccessible or irrelevant. It also requires a synchronized global state, which distributed sensors may lack because of communication delays, asynchronous updates, and stale remote data. Processing a smaller, locally reachable subset reduces these costs, but its RSO composition changes between decision epochs, continually remapping objects to fixed observation slots and action indices. The network must therefore learn both the tasking rule and the changing mapping between physical targets, inputs, and actions. Recurrent policies face the added difficulty of associating memory from previous slots with different objects at the current epoch.

Some scalability-oriented DRL approaches accommodate larger catalogues by projecting them onto angular grids \cite{siew2023scalable}. This representation merges RSOs that occupy the same cell, processes maps containing mostly empty locations, and assigns actions through grid cells rather than individual objects. Its computational footprint is therefore coupled to the spatial discretization rather than the number of decision-relevant entities. The policy also lacks recurrent memory beyond instantaneous motion features. This design philosophy represents a first step towards a scalable solution, but makes it computationally expensive and harder to preserve object identity, changing action semantics, and temporal context as targets move through the candidate set. Consequently, recent survey work accordingly identifies DRL based large-scale multi-sensor tasking as
an open problem in SSA \cite{xue2024review}. 

All these limitations motivated us to draw on more expressive neural architectures from other domains and develop a revised formulation for sensor tasking. Successful neural designs, including those used in AlphaStar \cite{vinyals2019grandmaster} and multi-agent autocurricula \cite{baker2019emergent}, provide a clear
architectural precedent: attention reasons over changing collections of
interacting entities \cite{NIPS2017_3f5ee243}, recurrence supports temporal
decision making \cite{hochreiter1997long}, and pointer decoding selects an action within a dynamic feasible set \cite{NIPS2015_29921001}. The same structure transfers
naturally to sensor tasking, where sensors and orbiting bodies can be represented as entity
tokens, allowing attention to compare their geometric, informational, and
cooperative relationships, a recurrent core carries temporal data, and a pointer selects through the current
candidate representations rather than fixed output labels. More recently, space-scheduling
studies have used pointer decoding for assignment and variable-cardinality
schedules \cite{liu2024enhancing,dobariya2026automatic}. However, the combination
studied in our research remains less explored: a flexible policy that decouples neural-network
size from catalogue size and sensor count, enabling the same architecture to
generalize across changing RSO populations and sensing configurations.

To address this gap, we introduce the Variable-Entity Intelligent Sensor Tasking Architecture (VISTA), which combines bounded entity-centric attention, a recurrent temporal core, and adaptive pointer selection over changing candidate sets. Each agent controls one sensor while encoding targets and teammates as entities, supporting heterogeneous-network coordination. Shared encoders preserve token-to-object correspondence and accommodate changing candidate and teammate counts. All agents jointly optimize the same distributed policy under a catalogue-level objective, then execute it from their own local observations. This follows the parameter-shared decentralized multi-agent paradigm used in large-scale deep reinforcement learning systems such as OpenAI Five \cite{berner2019dota} and in other sensor tasking approaches \cite{mignocchi2026meta,siew2023scalable}. In summary, our contributions are the following:
\begin{enumerate}
    \item a flexible neural architecture designed for the combinatorial and non-stationary nature of large-scale distributed sensor tasking, enabling generalization across changing catalogue sizes and uncertainty metrics while exhibiting a near-linear scaling law relating the number of sensors, the number of RSOs, and the tasking horizon;

    \item a configurable, physics- and mission-informed top-$K$ retrieval mechanism that focuses each agent on decision-relevant objects while bounding attention span and action space independently of catalogue size.

\end{enumerate}

\begin{figure*}[p]
    \centering
    \includegraphics[
        width=\textwidth,
        height=0.9\textheight,
        keepaspectratio
    ]{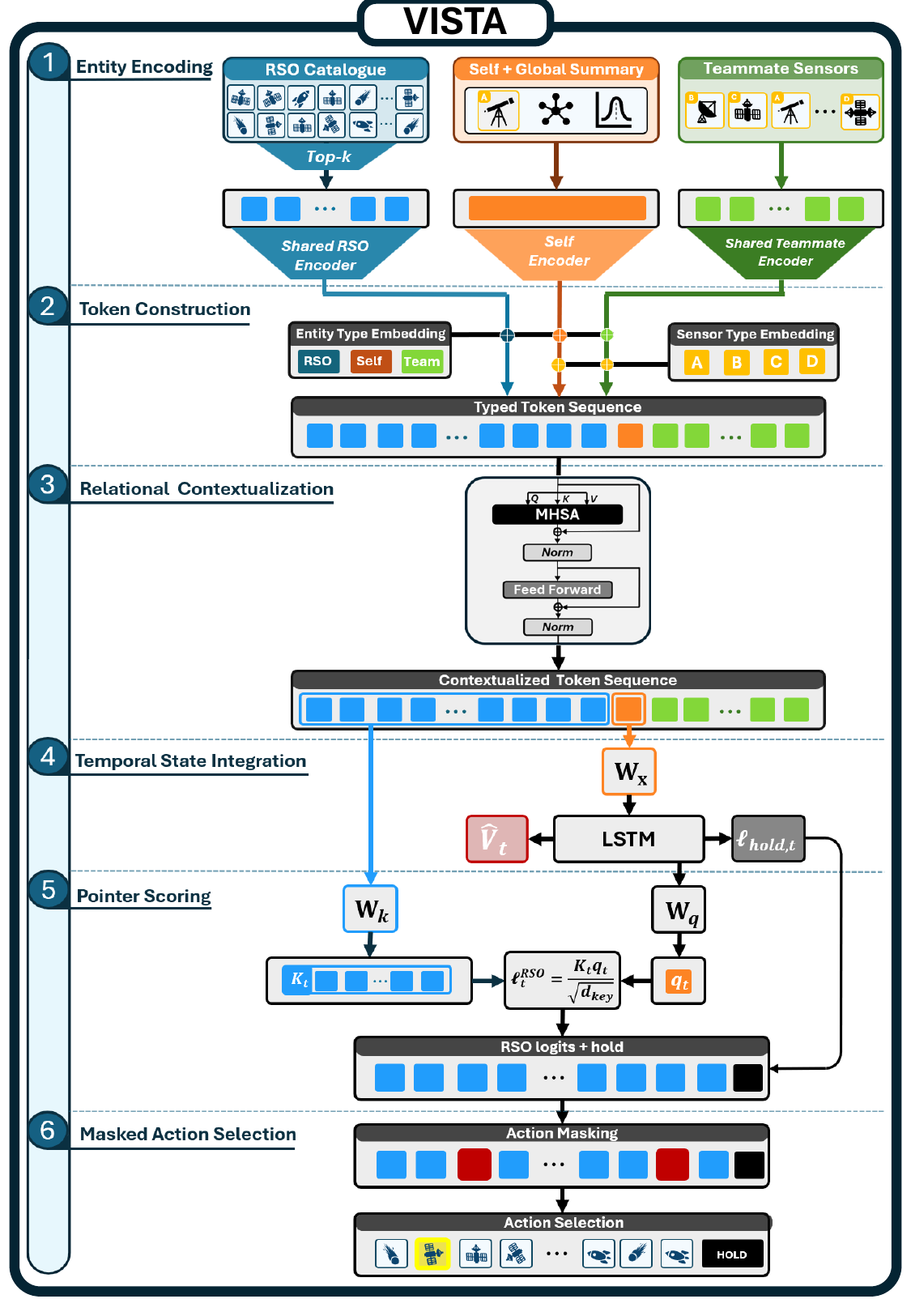}
    \caption{VISTA data path for each agent in charge of a sensor. Top-$K$ retrieval
    bounds the candidate set. Type-specific encoders and embeddings construct
    sensor and RSO tokens, self-attention models their relationships, and the
    LSTM integrates tasking history. Pointer scores and a hold logit form the
    masked action distribution, while the value head estimates state value.}
    \label{fig:pointer-scheme}
\end{figure*}

VISTA is evaluated in fixed-size, large-scale complex-motion, and heterogeneous
scenarios against heuristic, model-based, and recurrent baselines. It recovers
the 30-RSO catalogue 31.2\% faster than the recurrent baseline and reduces
five-hour large-scale uncertainty by 97.5\% relative to the strongest classical
reference. Tests with up to 20,000 objects reveal near-linear relationships
among sensing capacity, catalogue size, and recovery horizon. Heterogeneous-sensor
and representation analyses further show modality-adaptive tasking and
physically structured internal states.

\newpage

%% file: Sections/problem_formulation.tex
\subsection{Sensor Tasking Problem}

Consider a discrete-time sensor-tasking problem with decision epochs
$k = 0,1,\ldots,H-1$, where $H$ is the finite planning horizon. At each epoch,
the system maintains a catalogue $\mathcal{O}_k$ of resident space objects $o_{i,k}$,
\begin{equation}
\mathcal{O}_k =
\{o_{1,k}, o_{2,k}, \ldots, o_{N_k,k}\},
\end{equation}
where $N_k$ is the number of catalogue objects considered at epoch $k$. Each
object is represented by an estimated orbital state, an associated uncertainty
description, and additional tasking-relevant attributes. We write the catalogue
state compactly as
\begin{equation}
\mathcal{C}_k =
    \left\{
    \left(
    \hat{\mathbf{x}}_{i,k},
    \mathbf{P}_{i,k},
    \boldsymbol{\phi}_{i,k}
    \right)
    \right\}_{i=1}^{N_k}.
\end{equation}
where $\hat{\mathbf{x}}_{i,k}$ denotes the estimated state of object $i$,
$\mathbf{P}_{i,k}$ denotes its estimation uncertainty, and
$\boldsymbol{\phi}_{i,k}$ collects additional attributes such as priority,
object class, brightness, radar cross section, observation age, or
sensor-dependent observability features. 

The sensing network $S$ at epoch $k$ is
\begin{equation}
\mathcal{S}_k =
\{s_{1,k}, s_{2,k}, \ldots, s_{M_k,k}\},
\end{equation}
where $M_k$ is the number of available sensing assets. Each sensor is described
by a platform state and a set of operational parameters,
\begin{equation}
s_{j,k}
=
\left(
\hat{\mathbf{x}}^{s}_{j,k},
\boldsymbol{\psi}_{j,k}
\right),
\end{equation}
where $\hat{\mathbf{x}}^{s}_{j,k}$ denotes the sensor platform state and
$\boldsymbol{\psi}_{j,k}$ includes properties such as sensing modality,
boresight direction, field of view, field of regard, slew-rate limit, range or sensitivity limits, cadence, and availability. This formulation is agnostic to the sensing method and can represent heterogeneous networks of ground-based or space-based sensors using optical, radar, laser-ranging, or other sensing modalities.

At each decision epoch, the tasking policy assigns each available sensor either
to one catalogue object or to no observation. Let
\begin{equation}
a_{j,k} \in \{0,1,\ldots,N_k\}
\end{equation}
denote the action selected for sensor $j$, where $a_{j,k}=i$ means that sensor
$j$ is tasked to observe object $o_{i,k}$, and $a_{j,k}=0$ denotes an idle or
no-observation action. The joint tasking decision is
\begin{equation}
\mathbf{a}_k =
\left(
a_{1,k},
a_{2,k},
\ldots,
a_{M_k,k}
\right).
\end{equation}

It is important to state that not every assignment is realizable: visibility, geometry, sensor constraints, or mission specifications define a binary feasibility mask
\begin{equation}
m_{ji,k} =
\begin{cases}
1, & \text{if sensor $j$ can observe object $i$,}\\
0, & \text{otherwise.}
\end{cases}
\end{equation}
The assignment $a_{j,k}=i$ is admissible only if $m_{ji,k}=1$.

After a joint tasking decision is applied, the catalogue evolves according to the SSA system's orbital propagation, uncertainty growth, observation outcomes, missed detections, and measurement updates. This transition is written compactly as

\begin{equation}
\mathcal{C}_{k+1}
\sim
p
\left(
\mathcal{C}_{k+1}
\mid
\mathcal{C}_k,
\mathcal{S}_k,
\mathbf{a}_k
\right).
\end{equation}

The tasking objective is to maintain catalogue quality over the planning
horizon while respecting sensing and operational constraints. Let
\begin{equation}
J(\mathcal{C}_k)
\end{equation}
be a scalar catalogue-quality cost derived from the uncertainty of the tracked
objects, and let
\begin{equation}
U_{\mathrm{ctrl}}(\mathbf{a}_k)
\end{equation}
represent optional control, slew, resource, or coordination costs associated
with the joint action. The sensor-tasking problem can then be generally described as
\begin{equation}
\begin{aligned}
\min_{\pi}\quad
&
\mathbb{E}_{\pi}
\Bigg[
\sum_{k=0}^{H-1}
\Big(
J(\mathcal{C}_{k+1})
+
\lambda U_{\mathrm{ctrl}}(\mathbf{a}_k)
\Big)
\Bigg].
\end{aligned}
\label{eq:tasking_objective}
\end{equation}
subject to
\begin{equation}
\mathbf{a}_k
=
\pi(\mathcal{I}_k),
\qquad
\mathbf{a}_k
\in
\mathcal{A}^{\mathrm{feas}}_k
\left(
\mathcal{C}_k,
\mathcal{S}_k
\right),
\end{equation}
where $\mathcal{I}_k$ denotes the information available to the scheduler or
agent policy $\pi$ at epoch $k$. In a centralized scheduler,
$\mathcal{I}_k$ may include the full catalogue and sensing-network state. In a
decentralized multi-agent setting, each sensor may instead act from a local
observation or compressed candidate set.

This formulation highlights the central challenge of SSA sensor tasking: the
scheduler must make sequential assignment decisions over a dynamic,
constraint-dependent, and potentially large action space, while accounting for
the long-term effect of present observations on future catalogue uncertainty.

\subsection{Deep Reinforcement Learning Formulation}

The sensor-tasking process can first be represented in centralized form as a
partially observable Markov decision process (POMDP),
\begin{equation}
    \mathcal{P}^{\mathrm{cent}}
    =
    \left\langle
        \mathcal{X},
        \mathcal{A},
        P,
        \Omega,
        Z,
        R,
        \gamma
    \right\rangle.
\end{equation}
The state space $\mathcal{X}$ contains the complete catalogue and sensing
network, with
\begin{equation}
    \mathbf{x}_k
    =
    \left(
        \mathcal{C}_k,
        \mathcal{S}_k
    \right).
\end{equation}
The action space $\mathcal{A}$ contains the feasible joint assignments,
$P(\mathbf{x}_{k+1}\mid\mathbf{x}_k,\mathbf{a}_k)$ defines the state
transition, and $\Omega$ and $Z$ define the observation space and observation
kernel. The reward $R(\mathbf{x}_k,\mathbf{a}_k,\mathbf{x}_{k+1})$ expresses
the task-dependent catalogue-maintenance objective and may include operational
costs or shaping terms, where its concrete implementation is given later.
The factor $\gamma\in[0,1]$ discounts future rewards.

At epoch $k$, the scheduler receives
$\mathbf{y}_k\sim Z(\cdot\mid\mathbf{x}_k)$. If
$\mathbf{y}_k=\mathbf{x}_k$, the process is fully observable and reduces to an
MDP\@. Otherwise, the information state $\mathcal{I}_k$ may comprise the current
observation or an observation-action history,
\begin{equation}
\begin{aligned}
    \boldsymbol{\tau}_k^{\mathrm{cent}}
    &=
    \left(
        \mathbf{y}_0,
        \mathbf{a}_0,
        \ldots,
        \mathbf{y}_k
    \right), \\
    \mathbf{a}_k
    &\sim
    \pi^{\mathrm{cent}}
    \left(
        \cdot
        \mid
        \boldsymbol{\tau}_k^{\mathrm{cent}}
    \right).
\end{aligned}
\end{equation}
History dependence allows the policy to use temporal context when the current observation is
insufficient for decision-making. Timestamps, availability masks, or other metadata can also form part of
$\mathcal{I}_k$.

The centralized formulation extends naturally from the single-sensor case, but
its complexity grows rapidly with the number of sensors and targets. Before
feasibility constraints are applied, coordinating $M_k$ sensors over $N_k$
RSOs and an idle action yields up to $(N_k+1)^{M_k}$ possible joint
assignments. Centralized operation also requires a sufficiently current and
consistent view of the complete network, an increasingly restrictive
assumption for distributed sensors operating with delayed, intermittent, or
asynchronous communication.

For the regime in which all sensors share an identical total reward, the cooperative decentralized POMDP (Dec-POMDP) formulation is
\begin{equation}
    \mathcal{P}^{\mathrm{dec}}
    =
    \left\langle
        \mathcal{J},
        \mathcal{X},
        \{\mathcal{A}_j\}_{j\in\mathcal{J}},
        P,
        \{\Omega_j\}_{j\in\mathcal{J}},
        Z,
        R,
        \gamma
    \right\rangle ,
\end{equation}
where $\mathcal{J}$ indexes the sensing agents. The global state, transition,
and reward are unchanged, but the joint action is factorized across the local
action spaces and agent $j$ acts from its own information. Its observation and
history are
\begin{equation}
\begin{aligned}
    \mathbf{o}_{j,k}
    &= h_j(\mathbf{x}_k), \\
    \boldsymbol{\tau}_{j,k}
    &=
    \left(
        \mathbf{o}_{j,0},
        a_{j,0},
        \ldots,
        \mathbf{o}_{j,k}
    \right),
\end{aligned}
\end{equation}
where $h_j(\cdot)$ exposes the local sensor state, available coordination
information, and a bounded candidate subset of the catalogue. VISTA summarizes
this history recurrently, although the Dec-POMDP itself is agnostic to the
memory architecture.

Agents share policy parameters and, in the shared-reward regime, they also receive identical total rewards. The decentralized joint policy factorizes as
\begin{equation}
\begin{aligned}
    \pi_{\theta}
    \left(
        \mathbf{a}_k
        \mid
        \boldsymbol{\tau}_k
    \right)
    =
    \prod_{j=1}^{M_k}
    \pi_{\theta}
    \left(
        a_{j,k}
        \mid
        \boldsymbol{\tau}_{j,k}
    \right),
\end{aligned}
\end{equation}
with the feasibility mask restricting each local action. 

Finally, the shared policy is trained to maximize the expected discounted
return,
\begin{equation}
    \theta^{\star}
    =
    \arg\max_{\theta}
    \mathbb{E}_{\pi_{\theta}}
    \left[
        \sum_{k=0}^{H-1}
        \gamma^k
        R(\mathbf{x}_k,\mathbf{a}_k,\mathbf{x}_{k+1})
    \right].
    \label{eq:marl_objective}
\end{equation}
Policy parameters are optimized with proximal policy optimization (PPO)
\cite{schulman2017proximal}. We adopt this algorithm because simple PPO-based multi-agent methods have
been shown to match or outperform competitive off-policy algorithms in both
final return and sample efficiency across several cooperative multi-agent
benchmarks without requiring domain-specific modifications \cite{yu2022surprising}. 
\newpage
Its clipped actor objective is

\begin{equation}
\begin{aligned}
L^{\mathrm{CLIP}}(\theta)
=\mathbb{E}_{j,k}\Big[\min\big(&
\rho_{j,k}(\theta)\widehat{A}_{j,k},\\
&\operatorname{clip}(\rho_{j,k}(\theta),1-\epsilon,1+\epsilon)
\widehat{A}_{j,k}\big)\Big],
\end{aligned}
\end{equation}

where $\rho_{j,k}$ is the probability ratio between the updated and rollout
policies, $\epsilon$ is the clipping parameter, and $\widehat{A}_{j,k}$ is
estimated using generalized advantage estimation (GAE) \cite{schulman2015gae}. The complete PPO loss
also includes value-function regression and entropy regularization. The
concrete reward, parameter-sharing procedure, candidate construction, and
policy architecture are described in the following sections.

%% file: Sections/learning_framework.tex
\subsection{Experimental Tasking Environment}

The simulator is implemented in C and integrated with PufferLib for parallel rollout collection, high throughput and large-scale experimentation \cite{suarez2024pufferlib}. It preserves relative orbital motion, pointing history, geometric access, and observation-driven covariance reduction while abstracting the detection, association, and operational orbit-determination chain. Its uncertainty is a controlled tasking surrogate, not a prediction of realized catalogue accuracy. The principal physical settings are summarized in Table~\ref{tab:experiment-reproducibility}, while the complete configuration is retained in the project repository.

\paragraph{Orbital and Covariance Propagation.}
Each orbiting sensor and RSO follows an analytical unperturbed two-body trajectory. The implementation advances mean anomaly, solves Kepler's equation using an eccentricity-dependent iteration rule, and transforms into inertial coordinates. Ground sites remain fixed in the simulated nonrotating frame. Perturbations, terrestrial rotation, and manoeuvres are not modeled in these campaigns.

The native environment propagates a six-state covariance about these known trajectories rather than simulating a complete noisy state-estimation pipeline. Covariance prediction uses a second-order transition approximation and integrated white-acceleration process-noise blocks. Eligible observations produce sensor-quality-dependent position pseudo-measurement updates followed by a Joseph-form covariance update. The state geometry exposed to the scheduler follows the propagated trajectories, not independently uncertain estimated positions. The complete covariance and observation models are available in the project repository.

\paragraph{Sensor Geometry and Pointing Dynamics.}

When the agent selects an object to look at, the environment advances time and orbital positions, moves each boresight towards its selected target at bounded rate, tests post-motion observation conditions, updates covariance, and computes reward and subsequent observations. A target may require repeated selections over several steps before it enters the field of view, and eligibility at selection does not guarantee a later measurement.

Observation quality is configurable to match each experiment's objective.
The fixed-ground-sensor comparisons include range, off-boresight angle, and
object size, weighted by 0.65, 0.25, and 0.10, respectively, to represent
unequal sensing opportunities. The large-scale experiments instead isolate
pointing, scheduling, and scaling under relative motion by disabling the range
gate and assigning unit quality to every admissible observation. Measurement
availability remains binary, and geometry still governs access and pointing,
but measurement strength no longer depends on range or object properties.
Finally, the heterogeneous experiments test capability-dependent allocation
using complementary range responses: Type A is favoured nearby, Type B farther
away, and their qualities coincide at 1900~km. These controlled response models
allow different tasking questions to be studied within the same framework.

All three phases expose an explicit hold action in addition to the available
target-selection actions. Hold preserves the current boresight orientation and
commands no new target, while measurement availability remains governed by
the phase-specific sensing logic. Learned and classical methods use the same
hold convention and action interface within each experiment.

\paragraph{Bounded Candidate Construction.}

As discussed in the introduction, exposing the complete catalogue to every
sensor couples neural inference cost to population size, including objects
irrelevant to the current decision. VISTA instead presents each agent with
at most $K$ RSO tokens. This bounded context separates policy size from
catalogue size and provides a direct computation-quality tradeoff without
restricting the catalogue maintained by the environment.

As self-attention cost grows quadratically with the number of tokens, its
practical span is limited by the available inference budget. VISTA constructs
this bounded context in two stages: feasibility filtering followed by a
low-cost physics- and mission-informed ranking. First, the feasible set for
sensor $i$ at epoch $k$ is
\begin{equation}
\mathcal{F}_{i,k}
=
\left\{
j\in\{1,\ldots,N_k\}:
g_{ij,k}^{(\ell)}=1,
\quad
\forall \ell
\right\},
\end{equation}
where $g_{ij,k}^{(\ell)}$ is feasibility test $\ell$, such as visibility,
range, field of view, Earth occultation, or pointing reachability. Objects
outside $\mathcal{F}_{i,k}$ are inadmissible and removed before action
selection.

Each remaining sensor-RSO pair is then assigned the priority score
\begin{equation}
\begin{split}
p_{ij,k}={}&
\alpha_U\widetilde U_{j,k}
+\alpha_{\tau}\widetilde\tau_{j,k}
+\alpha_q\widetilde q_{ij,k}
+\alpha_m\widetilde m_{j,k}\\
&-\alpha_{\theta}\widetilde{\Delta\theta}_{ij,k}
-\alpha_c\widetilde c_{ij,k},
\end{split}
\end{equation}
where the terms respectively represent catalogue uncertainty, time since the
last observation, predicted observation quality, mission priority, retargeting
requirement, and coordination or redundancy. The marked quantities may be
normalized, clipped, log-scaled, or left unchanged according to the scenario.

The candidate set is then defined as
\begin{equation}
\mathcal{K}_{i,k}
=
\operatorname{TopK}_{j\in\mathcal{F}_{i,k}}
\left(p_{ij,k}\right),
\qquad
|\mathcal{K}_{i,k}|\leq K.
\end{equation}
If fewer than $K$ feasible objects are available, unused slots are padded and
masked before attention and action selection.

For the reported experiments, the active ranking uses only catalogue
uncertainty and time since the last observation,
\begin{equation}
p_{ij,k}
=
\alpha_U U_{j,k}
+
\alpha_{\tau}\tau_{j,k},
\qquad
\alpha_U=1,
\quad
\alpha_{\tau}=0.5,
\end{equation}
while the quality, mission-priority, retargeting, and coordination coefficients
are set to zero. These values act on the configured feature scales and are not
intended as universal nondimensional weights. This initial campaign uses
uncertainty and age as a simple retrieval prior, leaving feature selection and
weight optimization to future studies.

The ranking is a retrieval prior, not the tasking policy. The neural policy
still compares the retained RSOs with the sensor state and coordination
information, and may select any candidate or remain idle. Excluded objects can
re-enter the context as feasibility, geometry, age, or uncertainty changes,
while the simulator continues to propagate, update, and evaluate the complete
catalogue. Thus, $K$ bounds per-agent inference cost without fixing the size of
the underlying population. Its effect is evaluated empirically in
Section~\ref{sec:results}.

\subsection{Reward and Parameter-Shared Learning}

\paragraph{Cooperative Catalogue-Level Reward.}

For the distributed formulation in Section~\ref{sec:problem-formulation},
we use parameter-shared PPO with a catalogue-level team reward to support
stable, cooperative learning. This approach is competitive in cooperative
multi-agent problems \cite{yu2022surprising} and reduces to the single-agent
objective for one sensor.

Reward development began with approximations from earlier DRL-based SSA
tasking studies \cite{linares2016dynamic,linares2017autonomous,siew2022space}
and successive simulator trials. Purely differential rewards,
$J(\mathcal{C}_t)-J(\mathcal{C}_{t+1})$, measure immediate information gain
but yielded weak learning in large-scale scenarios as per-step changes
diminished in a well-maintained catalogue. We therefore adopted absolute
catalogue-health rewards, retaining an improvement term in the tail-aware
variants described below. For RSO $j$ and the complete catalogue,
\begin{equation}
\begin{aligned}
U_{j,t}
&=
\sigma_{R,j,t}
+
\sigma_{T,j,t}
+
\sigma_{N,j,t},\\
\bar U_t
&=
\frac{1}{N_{\mathrm{RSO}}}
\sum_{j=1}^{N_{\mathrm{RSO}}}
U_{j,t},
\end{aligned}
\end{equation}
where the sigmas are the radial, tangential, and normal marginal position
standard deviations, expressed in kilometres.

The large-scale cooperative experiments use the mean-level reward
\begin{equation}
r_t
=
\max\left(
1-\frac{\bar U_t}{U_{\mathrm{ref}}},
-1
\right),
\end{equation}
where $U_{\mathrm{ref}}$ is the initial catalogue mean uncertainty. Evaluated
over the complete catalogue and shared among all agents, this absolute reward
encourages uncertainty reduction and maintenance below the initial level,
preserving reward contrast after the largest covariance reductions.

When applied to small RSO populations with fixed ground sensors, this
large-scale formulation produced policies that neglected high-uncertainty
tail objects. We added reward-shaping terms to address this neglect, using
the tail-aware catalogue-badness function
\begin{equation}
B_t=
w_\mu\psi\!\left(\frac{\bar U_t}{u_\mu}\right)
+w_\tau\psi\!\left(\frac{U_t^{\mathrm{tail}}}{u_\tau}\right)
+w_M\psi_M\!\left(\frac{U_t^{\max}}{u_M}\right),
\end{equation}
where $U_t^{\mathrm{tail}}$ is the mean of the configured upper catalogue tail,
$U_t^{\max}=\max_j U_{j,t}$, and $\psi$ and $\psi_M$ are increasing badness
functions. With $B_{\mathrm{ref}}$ denoting the configured target badness, the
associated reward is
\begin{equation}
\begin{aligned}
r_{i,t}=\operatorname{clip}\big[{}&
\lambda_\Delta(B_{t-1}-B_t)
+\lambda_L(B_{\mathrm{ref}}-B_t)
+b_t^{\mathrm{fine}}\\
&+\alpha_{\mathrm{local}}g_t
-\alpha_{\mathrm{dup}}d_t
-\alpha_{\mathrm{ctrl}}\rho_{i,t}^{2},-1,1\big].
\end{aligned}
\end{equation}
Here, $g_t$ is shared local-information credit, $d_t$ measures duplicate
tasking, and $\rho_{i,t}$ is the normalized angular displacement commanded by
agent $i$. All terms are shared except the individual control cost, whose
coefficient is zero throughout.

The single-sensor experiments include a bounded fine-precision bonus. The
heterogeneous team instead emphasizes the upper tail and adds shared
local-information credit and a duplicate-tasking penalty, with the fine bonus
disabled. These phase-specific choices align the reward with each experiment's
objective. Table~\ref{tab:reward-parameters} provides reproducible numerical
settings, which can be retuned for other uncertainty scales or operating
regimes. Complete functions are available in the project repository.

\input{Tables/reward_parameters}

\paragraph{Shared Multi-Agent Policy.}

All sensors share policy and value parameters but maintain separate recurrent
states. Actor and value heads use the same local recurrent representation,
without a privileged global critic. A single simulator maintains the catalogue
and supplies current shared covariance and age information, with all other
sensors represented as teammate tokens. Decisions and updates are synchronized;
communication delays, losses, and bandwidth limits are not modeled. Asynchronous
policy execution is possible, but learning and coordination with unequal update
times or stale information remain to be evaluated.

\newpage

%% file: Tables/reward_parameters.tex
\begin{table}[!t]
\centering
\caption{Reward settings for single-sensor (S), large-scale (L), and
heterogeneous (H) experiments. Uncertainty scales are in kilometres.
A dash denotes an unused parameter, and zero denotes a disabled term.}
\label{tab:reward-parameters}
\footnotesize
\setlength{\tabcolsep}{2.2pt}
\renewcommand{\arraystretch}{1.05}
\begin{tabular}{@{}lccc@{}}
\toprule[0.8pt]
\textbf{Parameter} & \textbf{S} & \textbf{L} & \textbf{H}\\
\midrule
$(u_\mu,u_\tau,u_M)$ & $(0.8,2.5,3.5)$ & - & $(0.8,2.5,3.5)$\\
$(w_\mu,w_\tau,w_M)$ & $(.55,.30,.15)$ & $(1,0,0)$ & $(.50,.40,.10)$\\
Upper-tail fraction & 0.10 & - & 0.05\\
$(\lambda_\Delta,\lambda_L)$ & $(4,.05)$ & - & $(4,.05)$\\
$U_{\mathrm{ref}}$ & - & Reset mean & -\\
$\alpha_{\mathrm{local}}$ & 0 & 0 & 0.10\\
$\alpha_{\mathrm{dup}}$ & - & - & 0.05\\
$\alpha_{\mathrm{ctrl}}$ & 0 & 0 & 0\\
Team-sharing coefficient & 1 & 1 & 1\\
Reward bounds & $[-1,1]$ & $[-1,1]$ & $[-1,1]$\\
\midrule[0.3pt]
\multicolumn{4}{@{}l}{\textbf{Fine-precision terms}}\\
Maximum onset / transition & $4.5/3.8$ & - & -\\
Transition width & 0.25 & - & -\\
Maximum badness weight & 0.05 & 0 & 0\\
Maximum bonus scale & 0.80 & 0 & 0\\
Mean onset & 1.00 & - & -\\
Mean bonus scale & 0.60 & 0 & 0\\
\bottomrule[0.8pt]
\end{tabular}
\end{table}

%% file: Sections/neural_architecture.tex
\subsection{Design and Entity Encoding}

As discussed above, a high-value subset of RSOs is selected at each decision
epoch to avoid processing the full catalogue. The sensor-tasking policy thus
reasons over a changing heterogeneous set of ego, neighbouring-sensor, and
candidate-RSO information. This changing composition exposes a key weakness of
fixed-vector recurrent baselines: input and output positions change meaning as
candidates enter, leave, or are reordered. VISTA therefore uses a set-based
policy that accommodates a changing candidate set while preserving the
correspondence between observed RSOs and target-selection actions.

The architecture follows the data path shown in
Fig.~\ref{fig:pointer-scheme}. Type-specific encoders first map heterogeneous
entity features into a common latent space. Self-attention then models the
geometric, informational, and cooperative relationships within the current
observation, an LSTM retains information from previous decisions, and a pointer
head scores the candidate RSOs together with a hold action.

Let $m\in\{\mathrm{ego},\mathrm{nbr},\mathrm{rso}\}$ denote the entity type and
let $\mathbf{x}_{m,n,t}$ contain the continuous features of entity $n$ at epoch $t$. Each
type has its own encoder because the corresponding physical quantities differ,
while the encoder parameters are shared among all entities of the same type.
The ego and neighbouring-sensor entities correspond to ``Self'' and ``Team''
in the diagram, respectively. Each entity is encoded as
\begin{equation}
    \mathbf{h}^{(0)}_{m,n,t}
    =
    \operatorname{GELU}
    \left(
        \mathbf{W}_{m}\mathbf{x}_{m,n,t}
        +\mathbf{b}_{m}
    \right)
    +\mathbf{e}_{m}
    +\mathbf{e}^{\mathrm{mod}}_{m,n}.
\end{equation}
Here, $\mathbf{e}_{m}$ is the learned entity type embedding and
$\mathbf{e}^{\mathrm{mod}}_{m,n}$ is the sensor type embedding, used in
scenarios with heterogeneous sensor types to identify the sensing modality or
capability of a sensor. It is zero for homogeneous sensor teams and for all RSO
tokens. The encoded entities form the typed token sequence
\begin{equation}
    \mathbf{H}^{(0)}_t
    =
    \left[
        \mathbf{h}^{(0)}_{\mathrm{ego},t},
        \left\{\mathbf{h}^{(0)}_{\mathrm{nbr},j,t}\right\}_{j=1}^{N_{\mathrm{nbr},t}},
        \left\{\mathbf{h}^{(0)}_{\mathrm{rso},k,t}\right\}_{k\in\mathcal{K}_t}
    \right].
\end{equation}
No positional encoding is added, since token order carries no physical meaning.
When batching observations of different cardinalities, unused positions are
padded and masked. Table~\ref{tab:observation-tokens} summarizes the features
and their scaling, while Table~\ref{tab:experiment-reproducibility} gives
scenario-specific dimensions and settings. Full implementation details remain in
the project repository.

\input{Tables/observation_tokens}

\subsection{Attention and Recurrent Context}

Self-attention allows every entity to condition its representation on the other
entities currently visible to the policy. For attention head $g$, the query,
key, and value matrices are
\begin{equation}
\begin{aligned}
    \mathbf{Q}_{g,t} &= \mathbf{H}^{(0)}_t\mathbf{W}^{Q}_{g},\\
    \mathbf{K}_{g,t} &= \mathbf{H}^{(0)}_t\mathbf{W}^{K}_{g},\\
    \mathbf{V}_{g,t} &= \mathbf{H}^{(0)}_t\mathbf{W}^{V}_{g}.
\end{aligned}
\end{equation}
The corresponding contextual representation is
\begin{equation}
    \mathbf{C}_{g,t}
    =
    \operatorname{softmax}
    \left(
        \frac{
            \mathbf{Q}_{g,t}\mathbf{K}_{g,t}^{\mathsf T}
        }{
            \sqrt{d_g}
        }
        +\mathbf{M}_t
    \right)
    \mathbf{V}_{g,t},
\end{equation}
where $d_g$ is the key dimension and $\mathbf{M}_t$ masks padded tokens. The
heads are concatenated and projected following multihead self-attention,
\begin{equation}
    \operatorname{MHSA}
    \left(\mathbf{H}^{(0)}_t\right)
    =
    \operatorname{Concat}
    \left(
        \mathbf{C}_{1,t},\ldots,\mathbf{C}_{G,t}
    \right)
    \mathbf{W}^{O},
\end{equation}
and a transformer block combines this operation with residual connections,
normalization, and a position-wise feed-forward network. As the same
operations are applied to every token and no positional encoding is used,
reordering the entities produces the corresponding reordering of their
contextual representations, forming the contextualized token sequence in the
diagram.

Attention describes the current sensing situation but does not recover
information omitted from an instantaneous snapshot. VISTA therefore extracts the
contextualized ego token, projects it into the recurrent space, and updates an
LSTM state \cite{hochreiter1997long}:
\begin{equation}
\begin{aligned}
    \mathbf{z}_t
    &=
    \mathbf{W}_{\mathrm{rec}}
    \widetilde{\mathbf{h}}_{\mathrm{ego},t}
    +\mathbf{b}_{\mathrm{rec}},\\
    \left(
        \mathbf{h}^{\mathrm{rec}}_t,
        \mathbf{c}^{\mathrm{rec}}_t
    \right)
    &=
    \operatorname{LSTM}
    \left(
        \mathbf{z}_t,
        \mathbf{h}^{\mathrm{rec}}_{t-1},
        \mathbf{c}^{\mathrm{rec}}_{t-1}
    \right).
\end{aligned}
\end{equation}
The recurrent input projection $\mathbf{W}_{\mathrm{rec}}$ is labelled
$\mathbf{W}_x$ in the diagram. The recurrent state summarizes
decision-relevant temporal information, such as
recent assignments, pointing evolution, and changes in catalogue condition,
while the candidate representations retain their current entity-specific
context.

\subsection{Pointer and Value Heads}

The action decoder treats target selection as a comparison between the temporal
state of the controlled sensor and the contextual representation of each
candidate. The recurrent state produces a query, while the candidate RSO tokens
produce keys:
\begin{equation}
\begin{aligned}
    \mathbf{q}_t
    &=
    \mathbf{W}_{q}\mathbf{h}^{\mathrm{rec}}_t
    +\mathbf{b}_{q},\\
    \mathbf{k}_{k,t}
    &=
    \mathbf{W}_{k}\widetilde{\mathbf{h}}_{\mathrm{rso},k,t}
    +\mathbf{b}_{k},\\
    \ell_{k,t}
    &=
    \frac{
        \mathbf{q}_t^{\mathsf T}\mathbf{k}_{k,t}
    }{
        \sqrt{d_{\mathrm{key}}}
    },\\
    \ell_{\mathrm{hold},t}
    &=
    \mathbf{w}_{\mathrm{hold}}^{\mathsf T}
    \mathbf{h}^{\mathrm{rec}}_t
    +b_{\mathrm{hold}},\\
    \widehat{V}_t
    &=
    \mathbf{w}_{V}^{\mathsf T}
    \mathbf{h}^{\mathrm{rec}}_t
    +b_V.
\end{aligned}
\end{equation}
The shared candidate-key map avoids target-specific output parameters, while
the hold logit represents the option of retaining the current pointing command.
Let $\mathcal{A}_t$ contain the selectable candidate slots and the hold action.
The masked policy is a softmax over these valid logits,
\begin{equation}
    \pi_{\theta}
    \left(
        a_t=b\mid\boldsymbol{\tau}_t
    \right)
    =
    \left[\operatorname{softmax}
    \left((\ell_{c,t})_{c\in\mathcal{A}_t}\right)\right]_b,
    \qquad
    b\in\mathcal{A}_t,
\end{equation}
where infeasible or padded candidates are excluded before normalization.

The selected index identifies the RSO represented by the corresponding input
token rather than a permanent target label. Reordering the candidate tokens
therefore permutes their logits and probabilities, preserving the policy
distribution over physical targets. Unlike the original sequence-generating
Pointer Network, this adaptation produces one categorical action per decision
epoch, without autoregressive decoding within that epoch. Temporal recurrence
is instead provided by the LSTM.

%% file: Tables/observation_tokens.tex
\begin{table*}[t]
    \centering
    \caption{Entity-level observation layout used by the policy. Here, $K$ is
    the configurable top-$K$ candidate budget and $N_{\mathrm{nbr}}$ is the
    number of teammate sensors represented in each local observation. The
    base normalization constants are $d_0=10{,}000$~km, $v_0=10$~km/s,
    $u_0=10$~km, $u_{v,0}=10^{-3}$~km/s, and $\tau_0=3000$~s. All transforms
    are applied componentwise. The global catalogue summary is included in
    all three phases. H denotes the heterogeneous configuration.
    Configuration-specific extensions and scaling are retained in the repository.}
    \label{tab:observation-tokens}
    \footnotesize
    \setlength{\tabcolsep}{5pt}
    \renewcommand{\arraystretch}{1.05}
    \begin{tabular}{@{}>{\raggedright\arraybackslash}p{0.12\textwidth}>{\raggedright\arraybackslash}p{0.27\textwidth}>{\raggedright\arraybackslash}p{0.07\textwidth}>{\raggedright\arraybackslash}p{0.47\textwidth}@{}}
        \toprule
        \textbf{Token} & \textbf{Feature group} & \textbf{Dim.} &
        \textbf{Scaling and interpretation}\\
        \midrule
        Ego sensor
        & Position / velocity
        & $3+3$
        & $\tanh(\mathbf{r}/d_0)$ and $\tanh(\mathbf{v}/v_0)$\\
        & Boresight unit vector
        & $3$
        & Components in $[-1,1]$, using the local east-north-up frame in H\\
        & Previous pointing command
        & $2$
        & Normalized previous action slot and an auxiliary component\\
        & FoV half-angle
        & $1$
        & $\theta_{\mathrm{fov}}/\pi$\\
        & Modality (H only)
        & $1$
        & Identifier mapped to a learned embedding, not a continuous input\\
        & Range / slew capability (H only)
        & $1+1$
        & Maximum observation range and maximum slew rate\\
        & Global catalogue summary
        & $3$
        & Global mean and maximum uncertainty, and fraction of RSOs above
          the uncertainty threshold\\
        \midrule
        Teammate sensor
        & Relative position
        & $3$
        & $\tanh\!\left((\mathbf{r}_j-\mathbf{r}_i)/d_0\right)$\\
        & Boresight unit vector
        & $3$
        & Components already in $[-1,1]$\\
        & Previous-step sensing flag
        & $1$
        & Binary indicator that the teammate observed an RSO\\
        & Heterogeneous extension
        & $4$
        & Additional sensor descriptors, including modality, in H\\
        \midrule
        Candidate RSO
        & Relative position / velocity
        & $3+3$
        & $\tanh\!\left((\mathbf{r}_k-\mathbf{r}_i)/d_0\right)$ and
          $\tanh\!\left((\mathbf{v}_k-\mathbf{v}_i)/v_0\right)$\\
        & RTN position and velocity uncertainty
        & $3+1$
        & $\tanh[\log(1+\boldsymbol{\sigma}/u_0)]$ and
          $\tanh[\log(1+\sigma_v/u_{v,0})]$\\
        & Catalogue age
        & $1$
        & $\tanh[\log(1+\tau_k/\tau_0)]$\\
        & Object size
        & $1$
        & Characteristic size divided by $10$\\
        & Reachability / observation quality
        & $1+1$
        & Feasibility flag and a configuration-dependent quality or angular
          proximity feature, zero outside the current FoV\\
        & Local tasking history / coordination
        & $2+1$
        & Previous-target flag; clipped local revisit counter, with $-1$ for
          never observed; number of observing agents divided by team size\\
        & Relative azimuth / elevation
        & $2$
        & Offset divided by one-step slew capacity, clipped to $[-3,3]$, then
          divided by $3$\\
        \midrule
        Total
        & One ego sensor, $N_{\mathrm{nbr}}$ teammates, and $K$ candidate RSOs
        & $d_{\mathrm{obs}}$
        & Base: $15+7N_{\mathrm{nbr}}+19K$.
          H: $18+11N_{\mathrm{nbr}}+19K$, with modality IDs embedded separately\\
        \bottomrule
    \end{tabular}
    \renewcommand{\arraystretch}{1.0}
\end{table*}

%% file: Sections/experimental_setup.tex
The campaign examines fixed-size catalogue recovery, cooperative large-scale
sensing, and heterogeneous allocation under overlapping access. Each sensing
regime is trained independently, and the selected policy is then held fixed
during evaluation. All methods operate within the same covariance, geometry,
candidate, and action interfaces so that their tasking decisions can be
compared under common conditions. The complete environment and training
configuration is summarized in Table~\ref{tab:experiment-reproducibility},
with the full frozen files and supplementary records retained in the project
repository.

\input{Tables/experiment_reproducibility}

\subsection{Compared Methods and Interface Scope}
\label{sec:setup-methods}

The non-learning baselines form a progression of increasing scheduling
complexity. Random feasible selection provides an uninformed reference and
therefore quantifies the value of any task-dependent prioritization.
Oldest-first introduces a simple coverage and revisit strategy, while maximum
uncertainty (Max-$U$) greedily directs sensing resources towards the least
precisely maintained objects. Expected information gain (EIG) adds a
one-step, model-informed estimate of the uncertainty reduction produced by an
observation, approximated here by the sum of the predicted reductions in the
marginal position standard deviations. Together, these methods distinguish gains due to basic prioritization, catalogue-state
awareness, and observation-value prediction.

To assess whether short-horizon planning improves upon one-step selection, we
also include a lightweight receding-horizon beam search, deliberately simplified
relative to the Beam A* method of Federici et al.~\cite{federici2024optimal} to
remain computationally tractable in our large-scale experiments. It ranks the current candidates by EIG, expands a
width-four and depth-three beam, discourages duplicate assignments, and
executes the first action of the highest-scoring branch. Its rollout uses a
scalar uncertainty surrogate rather than repeatedly propagating the complete
future geometry, visibility, pointing dynamics, measurement quality, and
covariance update. This bounded surrogate preserves look-ahead planning within
the common online interface without the cost of full joint-state search.

The principal learned baseline is a fixed-dimensional recurrent policy based
on the MLP-LSTM formulation commonly used in earlier SSA tasking studies \cite{linares2016dynamic,linares2017autonomous,siew2022space,roberts2021deep,mignocchi2026meta}. It
uses fixed categorical target slots, providing a direct comparison with
VISTA's entity-centric encoding and pointer-based selection. Both policies use
the same PPO learning framework, task information, reward definition, and
environment interactions. However, each architecture receives its own model
configuration, hyperparameter optimization, rollout settings, and training
budget, allowing them to maximize their individual performance without being constrained by settings selected for the other network.

\subsection{Training and Evaluation Protocol}

Each learned architecture is trained through its own optimization campaign and
is evaluated only after its weights have been fixed. Training and evaluation
episodes span approximately $5.56$~h of simulated operation. Performance is measured using catalogue uncertainty,
where the uncertainty of each RSO is the sum of its radial, tangential, and
normal position standard deviations and the catalogue metric is their mean
across all RSOs. Recovery time is the first sampled instant at which this mean
reaches the specified precision threshold. The reduced single-sensor reference
uses 100 independently reset episodes. Computational cost limits the main
large-scale comparisons and context sweeps to 48 episodes per condition,
with smaller samples for the capacity grid and dedicated diagnostics. 

For the one-factor zero-shot sweeps, absolute terminal uncertainty is
supplemented with a difficulty-adjusted degradation metric. Let $U_{m,e}(x)$ be
the terminal mean uncertainty of method $m$ in paired episode $e$ under test
condition $x$, and let $x_{\mathrm{ID}}$ denote its in-distribution reference.
We define
\begin{equation}
\begin{aligned}
d_{m,e}(x)
&=
\frac{
U_{m,e}(x)/U_{m,e}(x_{\mathrm{ID}})
}{
U_{\mathrm{EIG},e}(x)/U_{\mathrm{EIG},e}(x_{\mathrm{ID}})
},\\
D_m(x)
&=
\exp\!\left[
\frac{1}{n}
\sum_{e=1}^{n}
\log d_{m,e}(x)
\right].
\end{aligned}
\label{eq:difficulty-adjusted-transfer}
\end{equation}
The first ratio measures each method's drift from its own in-distribution
performance, while the second expresses that drift relative to EIG under the
same shift. The geometric mean is used because uncertainty and degradation
factors span several orders of magnitude, allowing proportional improvements
and degradations to be treated symmetrically in log space. Thus, $D_m=1$
denotes the same multiplicative degradation as EIG, whereas $D_m<1$ indicates
greater robustness to the shift.

Cooperation is evaluated using pairwise nonduplication and effective
catalogue coverage. At each epoch with at least two active assignments, the
pairwise nonduplication rate is the fraction of unordered active-sensor pairs
assigned to different RSOs. To quantify how evenly successful observations are
distributed, let $p_r$ be the fraction assigned to RSO $r$. The normalized
entropy-based effective coverage is
\begin{equation}
\eta_{\mathrm{eff}}=
\frac{\exp\!\left(-\sum_r p_r\log p_r\right)}{N_{\mathrm{RSO}}},
\qquad 0\log 0:=0.
\label{eq:effective-coverage}
\end{equation}
Here, $0\log 0:=0$ is the standard continuous-limit convention, since
$\lim_{p\to 0^+}p\log p=0$. Thus, an RSO with no observations contributes zero
to the entropy sum. The metric equals one for uniform allocation and
$1/N_{\mathrm{RSO}}$ when all observations are concentrated on one object.

\paragraph{Representation diagnostics.}
Contextualized ego representations and recurrent states sampled during
evaluation are jointly projected across scenarios into two dimensions using
UMAP \cite{mcinnes2018umap}, separately for each phase and state type.
Scenario markers and catalogue-uncertainty colours indicate local similarities
in the encoded states and their evolution during recovery. These visualizations
help assess whether the policy organizes its internal states primarily by
scenario or by shared stages of catalogue recovery, providing context for
the observed transfer behaviour.

Projection-induced scenario overlap is assessed using each sample's 15 nearest
neighbours before and after projection. The intrusion rate is the percentage
of samples whose neighbours shift from a same-scenario to an other-scenario
majority, with the other-scenario fraction increasing by more than 0.25.
This check helps identify apparent cross-scenario mixing introduced by the
projection rather than already present in the original representations. It
measures a specific distortion, not overall projection fidelity.

Attention-feature associations are reported separately for each head using
Spearman correlation for continuous features and point-biserial correlation
(Pearson correlation with a binary indicator) for features such as visibility.
These associations reveal which measured task features covary with attention
weights and whether individual heads exhibit distinct patterns, helping
interpret how the policy distributes attention across the available context.
These diagnostics describe representational organization, not causal feature
attribution.

\subsection{Campaign Phases}

\begin{itemize}
\item \textbf{Phase I: fixed reference and controlled transfer.} The reference has one sensor, 30 RSOs, and persistent action-slot identities. One-factor transfer changes catalogue size, inclination, eccentricity, initial uncertainty, or measurement noise while retaining the trained policies. UMAP projections and attention associations supplement the behavioural comparisons with representation diagnostics.
\item \textbf{Phase II: cooperative LEO-to-LEO scheduling.} The reference uses 12 sensors, 2000 RSOs, and $K=60$. Frozen-policy shifts in catalogue length, candidate context, orbital geometry, and network size test scalability and transfer. An $N_{\mathrm{RSO}}$-to-$N_{\mathrm{sensor}}$ grid extending to 20,000 RSOs evaluates how the frozen neural policy scales zero-shot as catalogue and sensing-network sizes change, relating catalogue load and sensing capacity to the resulting uncertainty and recovery time. Same Phase-I internal representation assessments are applied.
\item \textbf{Phase III: heterogeneous allocation.} Two Type-A and two Type-B sensors observe 240 RSOs under overlapping access and complementary range-dependent responses. Allocation geometry, duplication, catalogue coverage, and modality interventions test whether the shared policy adapts its decisions to sensors with different capabilities.
\end{itemize}

%% file: Tables/experiment_reproducibility.tex
\begin{table*}[p]
\centering
\caption{Experiment reproducibility summary. Distances are in kilometres,
angles are in degrees except initial anomaly (radians), and $\mathcal{U}$ denotes a uniform distribution.
Within neural-network entries, V and F denote VISTA and the flat LSTM,
respectively. Detailed configuration files, complete sweep definitions, and
additional evaluation records are provided in the project repository.}
\label{tab:experiment-reproducibility}
\tablefont
\setlength{\tabcolsep}{3pt}
\renewcommand{\arraystretch}{1.0}
\begin{tabular}{@{}>{\raggedright\arraybackslash}p{0.245\textwidth}>{\raggedright\arraybackslash}p{0.235\textwidth}>{\raggedright\arraybackslash}p{0.235\textwidth}>{\raggedright\arraybackslash}p{0.235\textwidth}@{}}
\toprule[0.9pt]
\textbf{Setting} & \textbf{Phase I} & \textbf{Phase II} & \textbf{Phase III} \\
\midrule[0.5pt]
\multicolumn{4}{@{}l}{\tablefont\bfseries Environment configuration}\\[-1pt]
\cmidrule(lr){1-4}
\addlinespace[1pt]
\multicolumn{4}{@{}l}{\hspace{0.6em}\tablefont\bfseries\itshape Scale and timing}\\[-1pt]
Sensors / RSOs / $K$ & $1/30/30$ & $12/2000/60$ & $4/240/60$\\
Decision interval & 5~s & 5~s & 5~s\\
Episode length & 5.56~h & 5.56~h & 5.56~h\\
\addlinespace[2pt]
\multicolumn{4}{@{}l}{\hspace{0.6em}\tablefont\bfseries\itshape RSO orbital elements}\\[-1pt]
Semimajor axis & $\mathcal{U}[7000,7150]$ & $\mathcal{U}[6800,7200]$ & $\mathcal{U}[7000,7150]$\\
Eccentricity & $\mathcal{U}[0,0.005]$ & $\mathcal{U}[0,0.02]$ & $\mathcal{U}[0,0.005]$\\
Inclination & $\mathcal{U}[35,85]$ & $\mathcal{U}[0,100]$ & $\mathcal{U}[35,85]$\\
RAAN & $\mathcal{U}[-5,5]$ & $\mathcal{U}[0,360]$ & $\mathcal{U}[-14,14]$\\
Initial anomaly & $\mathcal{U}[0,2\pi)$ & $\mathcal{U}[0,2\pi)$ & $\mathcal{U}[0,2\pi)$\\
\addlinespace[2pt]
\multicolumn{4}{@{}l}{\hspace{0.6em}\tablefont\bfseries\itshape Initial catalogue state}\\[-1pt]
Position $\sigma_R/\sigma_T/\sigma_N$ (km) & $\mathcal U[1,2]/\mathcal U[3,6]/\mathcal U[1,2]$ & $\mathcal U[1,5]/\mathcal U[2,20]/\mathcal U[1,5]$ & $\mathcal U[1,2]/\mathcal U[3,6]/\mathcal U[1,2]$\\
Velocity $\sigma$ (km/s) & $\mathcal{U}[0.001,0.002]$ & $\mathcal{U}[0.001,0.005]$ & $\mathcal{U}[0.001,0.002]$\\
Observation age (s) & $\mathcal{U}[500,3000]$ & $\mathcal{U}[100,10000]$ & $\mathcal{U}[500,3000]$\\
\addlinespace[2pt]
\multicolumn{4}{@{}l}{\hspace{0.6em}\tablefont\bfseries\itshape Sensor model}\\[-1pt]
FoV half-angle & $6^\circ$ & $4^\circ$ & $6^\circ$\\
Maximum slew rate & $0.5^\circ$/s & $5^\circ$/s & $0.5^\circ$/s\\
Range gate & 3000 & Disabled & 3200\\
Nominal position noise & 0.05 & 0.05 & 0.08\\
Quality model & Geometric & Geometry-agnostic & Complementary\\
\specialrule{0.8pt}{4pt}{1.5pt}
\multicolumn{4}{@{}l}{\tablefont\bfseries Neural network}\\[-1pt]
\cmidrule(lr){1-4}
\addlinespace[1pt]
\multicolumn{4}{@{}l}{\hspace{0.6em}\tablefont\bfseries\itshape VISTA}\\[-1pt]
Encoder input dimensions & Ego 15; RSO 19 & Ego 15; teammate 7; RSO 19 & Ego 18; teammate 11; RSO 19; modality-aware\\
Attention & 4 heads; width 32; FF 128 & 4 heads; width 64; FF 256 & 4 heads; width 64; FF 256\\
Recurrent core / output & LSTM 64; pointer and value & LSTM 64; pointer and value & LSTM 64; pointer and value\\
\addlinespace[2pt]
\multicolumn{4}{@{}l}{\hspace{0.6em}\tablefont\bfseries\itshape Flat LSTM}\\[-1pt]
Encoder & GELU MLP $582\!\to\!512\!\to\!512$ & GELU MLP $1229\!\to\!512\!\to\!512$ & -\\
Recurrent core / output & LSTM 512; categorical and value & LSTM 512; categorical and value & -\\
\addlinespace[2pt]
Total trainable parameters & V: 52,770; F: 2,678,816 & V: 98,625; F: 3,024,957 & V: 99,330\\
\specialrule{0.8pt}{4pt}{1.5pt}
\multicolumn{4}{@{}l}{\tablefont\bfseries Hyperparameter sweep}\\[-1pt]
\cmidrule(lr){1-4}
\addlinespace[1pt]
Campaign & VISTA and flat LSTM independently; PufferLib Protein & VISTA and flat LSTM independently; PufferLib Protein & No new sweep; Phase-I settings retained\\
Trial budget / metric & 250 million transitions; maximum catalogue uncertainty & 250 million transitions; mean catalogue uncertainty & -\\
Search ranges & Horizon $[16,256]$; entropy $[10^{-4},10^{-2}]$; $\gamma[0.5,0.99]$; LR $[3\!\times\!10^{-5},3\!\times\!10^{-2}]$ & Horizon $[16,256]$; entropy $[10^{-4},10^{-2}]$; $\gamma[0.5,0.999]$; LR $[3\!\times\!10^{-5},3\!\times\!10^{-2}]$ & -\\
\specialrule{0.8pt}{4pt}{1.5pt}
\multicolumn{4}{@{}l}{\tablefont\bfseries Optimized training parameters}\\[-1pt]
\cmidrule(lr){1-4}
\addlinespace[1pt]
Total transitions & V: 25 million; F: 700 million & V: 100 million; F: 150,183,936 & V: 25 million\\
Minibatch / batch & V/F: 16,384; automatic & V/F: 16,384; automatic & V: 16,384; automatic\\
Parallelization & $48\!\times\!256$ environments; 1 agent & $48\!\times\!12$ environments; 12 agents & $48\!\times\!256$ environments; 4 agents\\
Initial LR / schedule & V: $1.2\!\times\!10^{-3}$; F: $1.368322\!\times\!10^{-3}$; annealed & V: $7.019435\!\times\!10^{-4}$; F: $5\!\times\!10^{-3}$; constant & V: $1.2\!\times\!10^{-3}$; annealed\\
Entropy coefficient & V: $4\!\times\!10^{-4}$; F: $3.235870\!\times\!10^{-4}$ & V: $3\!\times\!10^{-4}$; F: $1.5\!\times\!10^{-4}$ & V: $4\!\times\!10^{-4}$\\
Discount / GAE $(\gamma,\lambda)$ & V: $(0.67,0.99)$; F: $(0.5101153,0.99)$ & V: $(0.67,0.99)$; F: $(0.5,0.999)$ & V: $(0.67,0.99)$\\
Horizon / epochs & V/F: 128 / 1 & V: 64 / 1; F: 32 / 1 & V: 128 / 1\\
Remaining PPO settings & PufferLib default & PufferLib default & PufferLib default\\
Reward (Table~\ref{tab:reward-parameters}) & Tail-aware catalogue badness with fine-precision terms & Shared mean level relative to reset uncertainty & Tail-aware catalogue badness with cooperative terms\\
\bottomrule[0.9pt]
\end{tabular}
\par\smallskip
\parbox{0.96\textwidth}{\tabnotefont
Experiment configurations available at
\url{https://github.com/RocketNeurons/VISTA-SSA}.
\par Compute server: Intel Xeon Gold 6430 (64 vCPUs, hyperthreading disabled),
one NVIDIA L40 GPU (48~GB), and 251~GiB system RAM.}
\renewcommand{\arraystretch}{1.0}
\end{table*}

%% file: Sections/results_and_discussion.tex
The following results characterize relative tasking performance within the
adopted orbital, sensing, and covariance models. Their absolute uncertainty
values should therefore be interpreted as controlled simulation outcomes rather
than predictions of operational catalogue accuracy. The complete sweep ranges,
policy dimensions, training hyperparameters, evaluation settings, and
supplementary numerical results are reported in
Table~\ref{tab:experiment-reproducibility} and the project repository.

\subsection{Phase I: Single Sensor with Small RSO Population}
\label{sec:results-phase1}
\subsubsection{In-Distribution Comparison}

\begin{figure*}[!t]
\centering
\includegraphics[width=\textwidth]{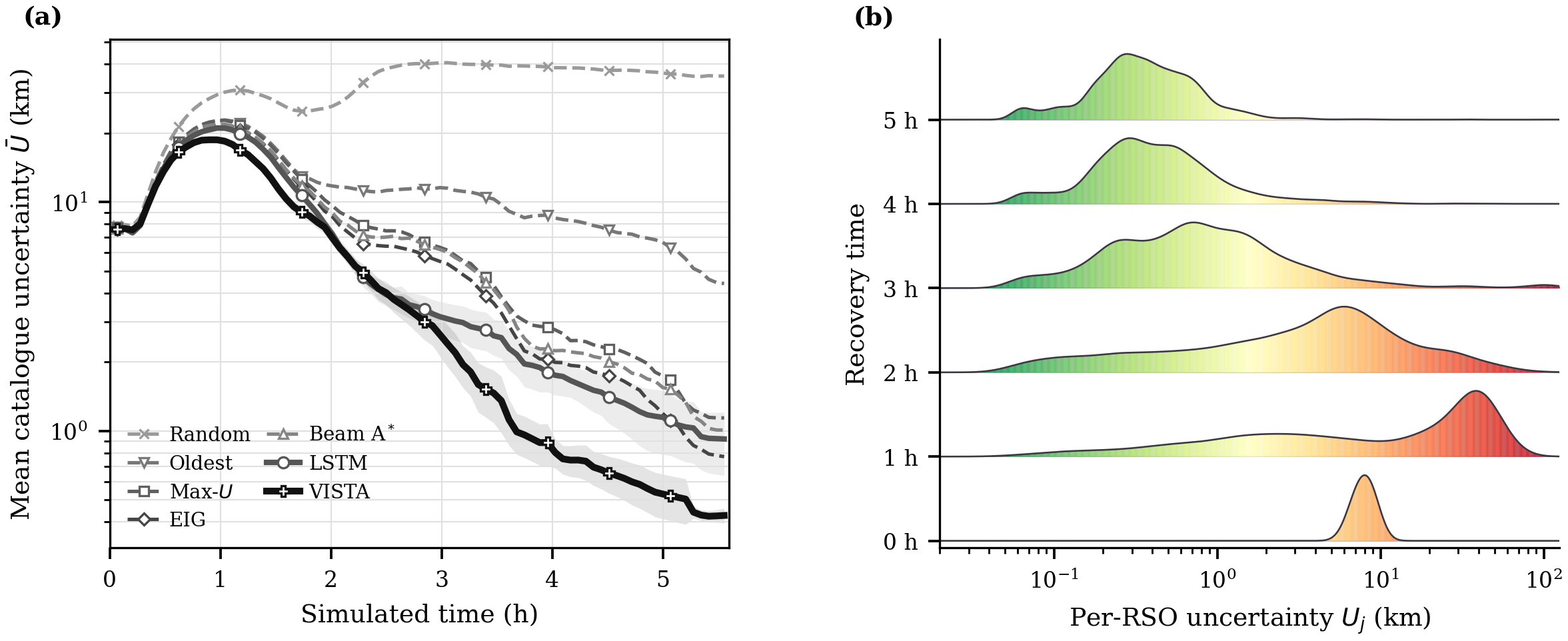}
\caption{Phase-I recovery with one sensor and 30 RSOs over 5.56~h.
\textbf{(a)} Mean catalogue uncertainty; shading shows 95\% confidence
intervals for the learned-policy mean trajectories across evaluation episodes.
\textbf{(b)} Hourly per-RSO uncertainty distributions for VISTA. Each ridge
is normalized independently for display.}
\label{fig:phase1-id}
\end{figure*}

Figure~\ref{fig:phase1-id}(a) shows the recovery achieved over the
5.56-h campaign. Across 100 evaluation episodes, VISTA reaches a terminal
mean uncertainty of 0.427~km, with a maximum of 1.861~km and a $p_{99}$ of
1.621~km. These values are 44.6\%, 55.6\%, and 56.0\% below EIG,
respectively. Its time-averaged mean uncertainty is 5.814~km, a 23.4\%
reduction relative to EIG. The aggregate mean curve reaches 1~km after
13,245~s, compared with 18,745~s for EIG and 19,245~s for the flat LSTM,
representing reductions of 29.3\% and 31.2\% in recovery time. VISTA is also
the only method whose aggregate curve reaches 0.5~km within the evaluation
horizon.

The comparison illustrates the progression among the non-learning schedulers.
Oldest-first and Max-$U$ prioritize observation age and uncertainty, EIG adds
an estimate of immediate covariance reduction, and the surrogate beam search
introduces a short scalar rollout. The latter does not improve upon EIG in
this regime, while already requiring nearly 20 times the computation of a
VISTA decision. Wider or deeper search with full covariance and geometry
propagation would further increase this cost, particularly in the larger
experiments. This comparison therefore concerns the bounded surrogate, not
look-ahead planning in general.

The flat LSTM initially improves but later plateaus, whereas VISTA continues
reducing catalogue uncertainty. Despite having 2,678,816 parameters versus
VISTA's 52,770 and undergoing a 500-run architecture and hyperparameter search,
the flat baseline required longer runs and more of its sweep to reach its best
result. The 25-million- and 700-million-transition budgets for VISTA and the
flat LSTM reflect PufferLib's throughput-oriented, single-epoch PPO, which
favours stable near-on-policy updates over rollout reuse
\cite{suarez2024pufferlib}. Raw counts are therefore not comparable with
multi-epoch PPO budgets and indicate easier optimization rather than a
controlled sample-efficiency advantage.

Figure~\ref{fig:phase1-id}(b) shows the complete catalogue distribution rather
than only its mean. Each ridge represents the distribution of per-RSO
uncertainty at one hourly snapshot. The distribution first broadens as some
unobserved objects develop large uncertainty peaks. As tasking progresses,
VISTA dynamically revisits these poorly maintained objects, suppresses the
upper tail, and moves an increasing fraction of the catalogue towards the
low-uncertainty side. This evolution shows how the policy balances peak
reduction with broad catalogue recovery.

\subsubsection{Controlled Zero-Shot Generalization}
\label{sec:results-phase1-transfer}

\begin{figure*}[!t]
\centering
\includegraphics[width=\textwidth]{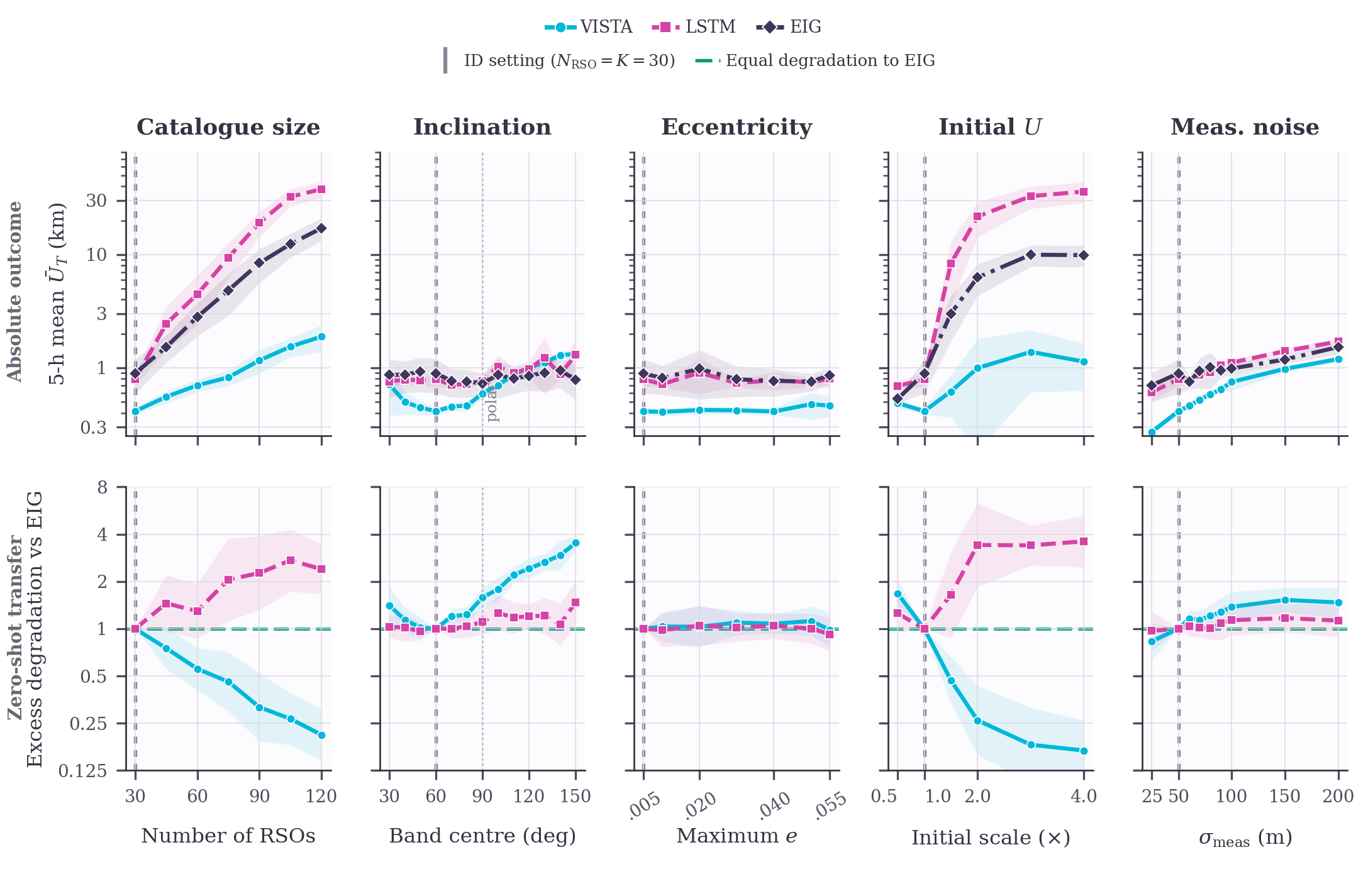}
\caption{Phase-I one-factor zero-shot transfer using 24 paired episodes per
condition. Upper panels show five-hour mean catalogue uncertainty. Lower panels
show the paired relative degradation $D_m$ defined in
Eq.~\eqref{eq:difficulty-adjusted-transfer}. Shading denotes 95\% confidence
intervals across episodes; lower-row intervals are computed from paired
log-ratios. Dashed grey vertical lines mark the in-distribution training
conditions, while the green horizontal line at $D_m=1$ denotes degradation
equal to EIG. Both rows should be interpreted jointly.}
\label{fig:phase1-transfer}
\end{figure*}

The upper panels in Fig.~\ref{fig:phase1-transfer} report zero-shot five-hour mean uncertainty, while the lower panels use the difficulty-adjusted degradation $D_m$ defined in Eq.~\eqref{eq:difficulty-adjusted-transfer}. This second metric distinguishes absolute tasking quality from the proportional drift caused by each
distribution shift. Both rows must therefore be assessed together: a method
with poor in-distribution performance may appear robust simply because it has
little additional performance to lose, whereas a stronger method may degrade
more proportionally while still achieving a substantially better absolute
outcome. Consequently, a lower $D_m$ alone does not imply a better scheduler.

At 120 RSOs with $K=30$, VISTA's terminal uncertainty is 89.0\% below EIG and
95.0\% below the flat LSTM, with $D=0.211$ compared with 2.398 for the LSTM.
When the initial uncertainty is multiplied by four, VISTA is 88.5\% below EIG
and 96.9\% below the LSTM, with $D=0.168$ compared with 3.606. These results
show that the entity-centric policy accommodates substantial changes in
catalogue population and initial condition without retraining.

Eccentricity transfer is comparatively stable, with an endpoint increase of
approximately 11.9\% between $e_{\max}=0.005$ and 0.055. At 200~m measurement
noise, VISTA reduces terminal uncertainty by 21.6\% relative to EIG and 30.1\%
relative to the LSTM. Its value $D=1.470$ nevertheless indicates greater
proportional degradation than EIG, illustrating that absolute performance and
relative sensitivity describe complementary aspects of transfer.

The inclination sweep produces the most revealing transfer result. Each test
samples RSO inclinations from a fixed $50^\circ$ wide band, and the reported
band centre specifies the midpoint of that distribution. Moving the centre to
$150^\circ$ therefore replaces the nominal orbital population with a strongly
different relative-motion and access geometry. Under this shift, VISTA becomes
2.4\% worse than the LSTM and 68.8\% worse than EIG, with $D=3.523$ compared
with 1.476 for the LSTM. This sharp and structured response is important: it
shows that VISTA is not merely fitting catalogue uncertainty statistics, but is
organizing its decisions around learned orbital relationships. That geometric
representation may be the source of its nominal strength, while also revealing that training on a single inclination regime can limit zero-shot transfer. Broader
geometric randomization during training is therefore a direct path towards
retaining this relational advantage across more diverse orbital populations.
\newpage

\subsubsection{Representation and Attention Diagnostics}

\begin{figure*}[!p]
\centering
\includegraphics[width=\textwidth,height=0.9\textheight,keepaspectratio]{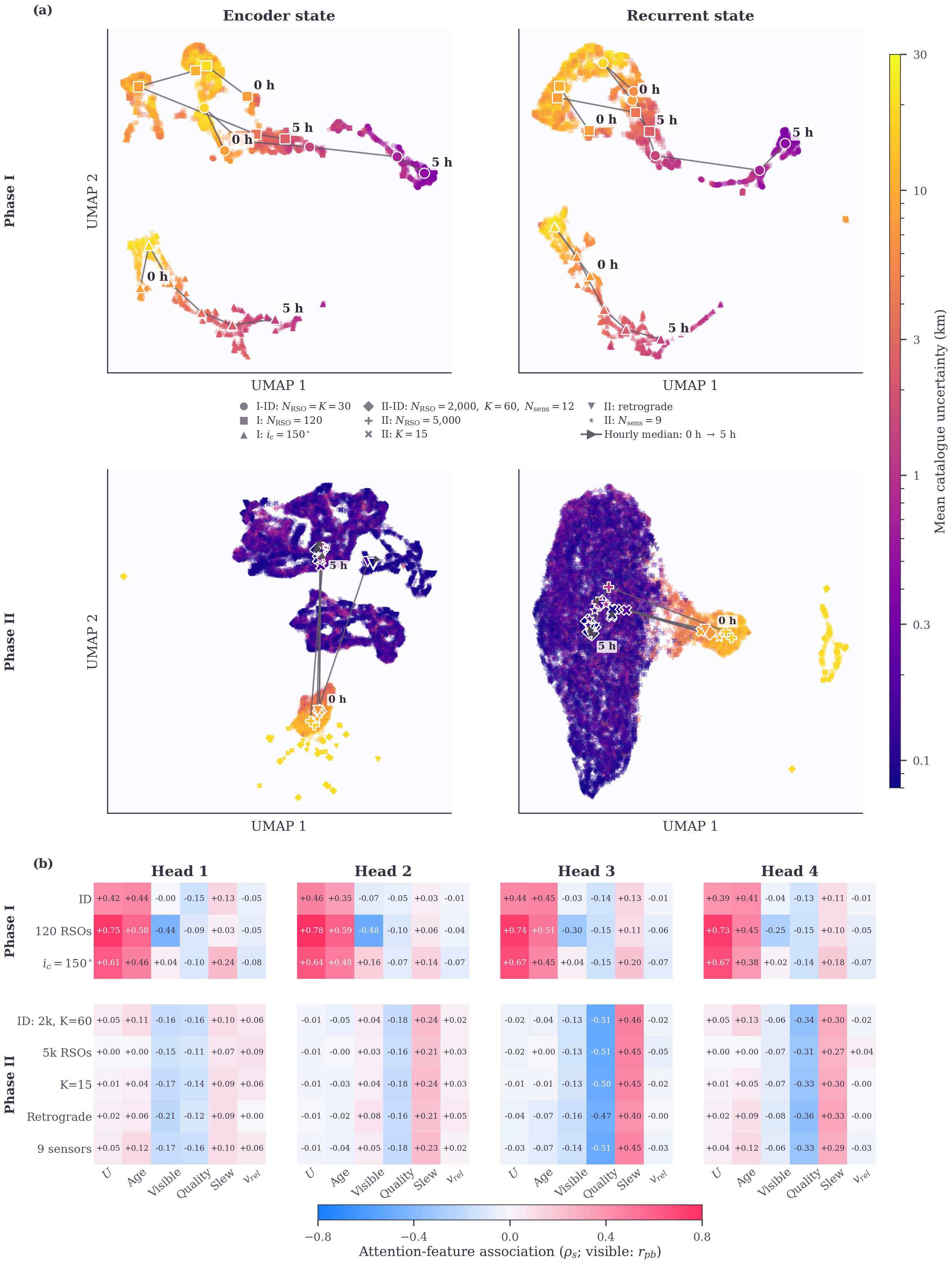}
\caption{VISTA representation and attention diagnostics, with Phase I above
Phase II in both blocks. \textbf{(a)} Joint UMAP projections of the encoder
summary entering the LSTM (left) and the recurrent state (right). Marker shapes
identify evaluation scenarios, colour denotes mean catalogue uncertainty on a
logarithmic scale, and arrows trace the temporal progression of the hourly
medians. Each state and phase uses a separate UMAP fit; absolute coordinates
are therefore not comparable between panels. \textbf{(b)} Associations between
ego-query attention scores and RSO-token features, with one column per
attention head. Continuous features use Spearman correlation $\rho_s$, whereas
visibility uses point-biserial correlation $r_{pb}$. These associations are
exploratory and do not establish causal feature importance.}
\label{fig:representation-diagnostics}
\end{figure*}

The in-distribution and larger-catalogue trajectories in
the upper row of Fig.~\ref{fig:representation-diagnostics}(a) largely occupy
the same manifold in both the encoder and recurrent projections and follow a
similar progression as catalogue uncertainty decreases, which is consistent
with their strong zero-shot transfer. The inclination samples instead form a
distinct branch, matching the performance loss observed when the orbital
geometry changes. Projection-intrusion rates of approximately 2.0\% to 4.25\%
for the reference and population-shift samples suggest limited projection-induced
scenario mixing, without validating every feature of the visible manifold.

The Phase-I heads in Fig.~\ref{fig:representation-diagnostics}(b) show similar
patterns, dominated by positive associations with uncertainty and observation
age. This is consistent with prioritizing poorly maintained objects among the
few available opportunities in a sparse perceptive field. Weaker associations
with slew suggest sensitivity to pointing requirements when several targets
are available. Together, these patterns are compatible with balancing
catalogue priority and economical retargeting, but attention alone does not
establish a particular visiting order or shortest-path strategy.
\newpage
\subsection{Phase II: Large-Scale Cooperative LEO-to-LEO Scheduling}
\label{sec:results-phase2}

\begin{figure*}[!t]
\centering
\includegraphics[width=\textwidth]{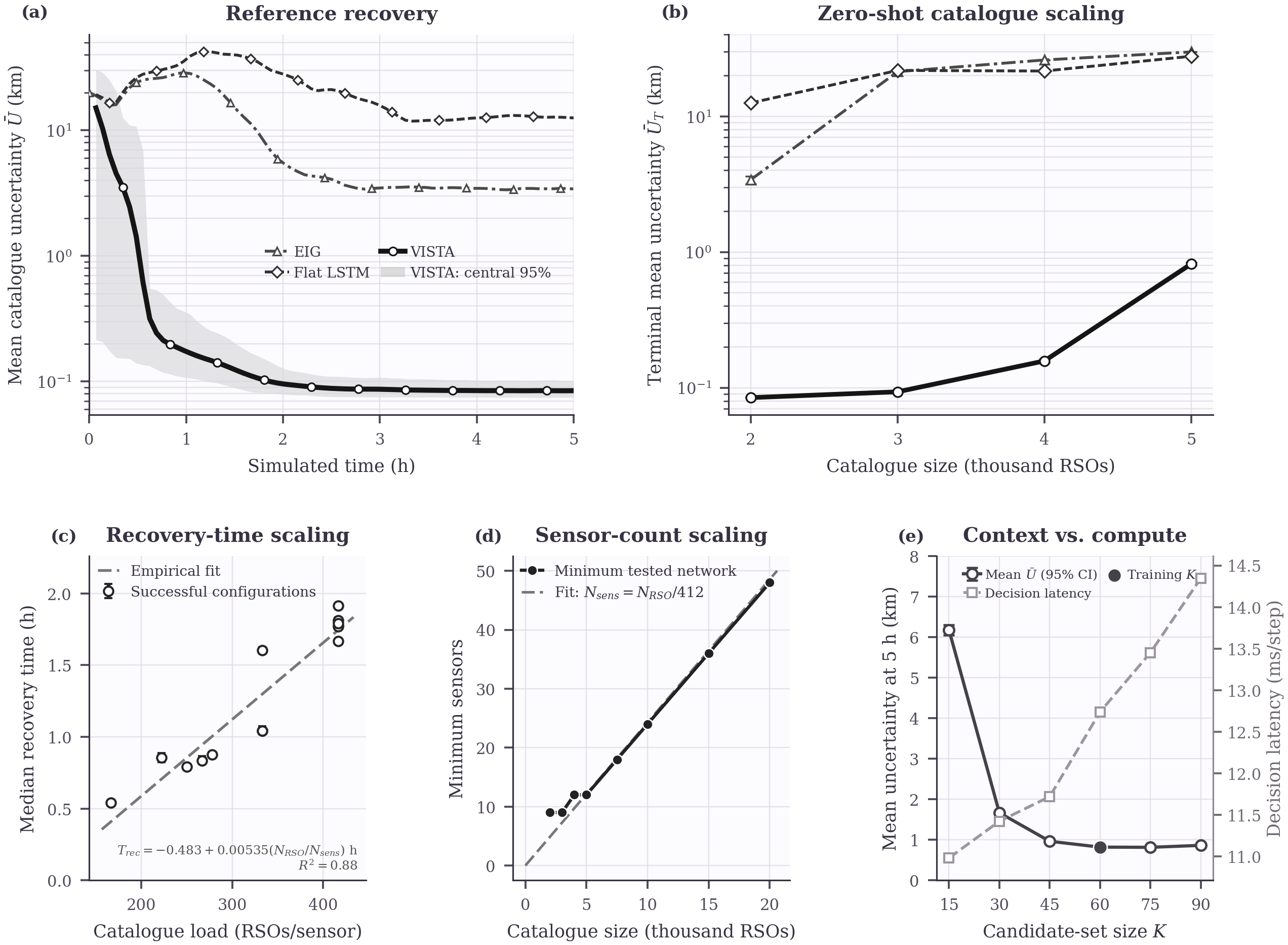}
\caption{Phase-II cooperative performance and frozen-policy scaling.
\textbf{(a)} Five-hour recovery with 12 sensors and 2000 RSOs. Shading shows
the central 95\% of VISTA's per-RSO uncertainty distribution, not a confidence
interval. \textbf{(b)} Five-hour mean uncertainty for 2000-5000 RSOs with
12 sensors; error bars show 95\% confidence intervals over 48 episodes.
For panels \textbf{(c)} and \textbf{(d)}, a configuration is successful when
at least 80\% of its episodes reach a mean catalogue uncertainty of 1~km
within five hours. \textbf{(c)} Median first-crossing time for 13 successful
configurations, with interquartile bars and a least-squares fit.
\textbf{(d)} Smallest successful tested sensor constellation and a
through-origin fit over eight catalogue sizes. \textbf{(e)} Five-hour mean
uncertainty and policy-call latency for batches of 144 agents as the candidate
set size $K$ varies. Uncertainty error bars show 95\% confidence intervals over
48 episodes, and the filled marker denotes the training value $K=60$. The
complete capacity grid and episode counts are provided in the project
repository.}
\label{fig:phase2-aggregated}
\end{figure*}

In the large-scale reference scenario with 12 orbital sensors and 2000 RSOs,
Fig.~\ref{fig:phase2-aggregated}(a) shows that VISTA reduces mean catalogue uncertainty
by 99.1\% after one hour and 99.5\% after two hours, before approaching a stable
low-uncertainty regime. At the five-hour endpoint, its mean uncertainty is
0.0846~km, compared with 3.415~km for EIG and 12.567~km for the flat LSTM. This
corresponds to reductions of 97.5\% and 99.3\%, respectively. The terminal
$p_{99}$ remains only approximately 29\% above the median, indicating that the
recovery is broadly distributed across the catalogue rather than concentrated
on a small subset of RSOs.

The faster reduction observed here than in Phase I should not be attributed to
the larger cooperative sensor network alone. Phase II also changes the orbital
geometry, increases the slew rate from $0.5^\circ$/s to $5^\circ$/s, disables
the range gate, and applies a unit-quality measurement model to every
admissible observation. Its recovery rate therefore reflects the combined
effect of increased sensing capacity, faster pointing, broader access, and a
geometry-agnostic measurement update.

\subsubsection{Zero-Shot Catalogue Scaling}

Figure~\ref{fig:phase2-aggregated}(b) evaluates the policies trained at 2000 RSOs
without further optimization. VISTA obtains terminal uncertainties of 0.0933,
0.1577, and 0.8183~km at 3000, 4000, and 5000 RSOs. The corresponding EIG values
are 21.358, 26.055, and 29.873~km, while the flat LSTM reaches 21.807, 21.591,
and 27.738~km. Across the complete 2000-5000-RSO sweep, VISTA is therefore
97.3-99.6\% below EIG and 97.0-99.6\% below the flat LSTM. Although VISTA's own
uncertainty increases as the number of objects per sensor grows, it remains
below 1~km at 5000 RSOs and preserves a substantial advantage over both
comparators. A fixed network has more objects to maintain as population grows,
so recovery takes longer and a common five-hour endpoint samples different
stages of that process. The rising endpoints therefore reflect increased
catalogue load as well as transfer sensitivity, not an isolated loss of
generalization.

\subsubsection{Empirical Recovery and Network Scaling}
For this dimensioning study, recovery is the first sampled time at which mean
catalogue uncertainty falls to 1~km or below. This common simulation target
measures how quickly a network restores catalogue quality, allowing sensor
count, catalogue load, and recovery horizon to be related without implying an
operational accuracy requirement. Figure~\ref{fig:phase2-aggregated}(c) shows
the 13 successful population-network configurations from a grid evaluated with
two, four, or six episodes per cell. Unweighted least squares gives
\begin{equation}
T_{\mathrm{rec}}[\mathrm{h}]=-0.483+0.00535\frac{N_{\mathrm{RSO}}}{N_{\mathrm{sensors}}},
\qquad R^2=0.882.
\label{eq:recovery-scaling}
\end{equation}
Thus, recovery time is strongly related to catalogue load per sensor within
the tested regime. The simple fit supports empirical network dimensioning as
population and the uncertainty distribution presented to top-$K$ retrieval
change, rather than requiring a separate policy for every network size.

The minimum tested networks in Fig.~\ref{fig:phase2-aggregated}(d) yield a
separate through-origin fit of
$N_{\mathrm{sensors}}\simeq N_{\mathrm{RSO}}/411.859$. Restricting the fit to
populations of at least 5000 RSOs gives one sensor per 416.667 RSOs, showing a
consistent near-linear trend over the large-population portion of the grid.
For example, the smallest successful tested networks contain 12, 18, 24, 36,
and 48 sensors for 5000, 7500, 10,000, 15,000, and 20,000 RSOs, respectively.

These relations are empirical scaling summaries rather than universal capacity
laws. At 20,000 RSOs, 42 sensors do not reach the threshold while 48 sensors do,
and intermediate network sizes were not tested. Moreover, resizing the network
changes constellation phasing as well as the per-sensor catalogue load. Further
trials and orbital geometries would therefore be required to establish the
generality of the fitted relationships.

\subsubsection{Candidate-Context Sensitivity}
The candidate budget $K$ is central to VISTA's scalability, but limiting
context risks excluding valuable targets or relationships. The sweep in
Fig.~\ref{fig:phase2-aggregated}(e), at 5000 RSOs, tests whether a small
high-priority subset suffices or whether additional context materially improves
tasking. Relative to $K=15$, terminal uncertainty decreases by 73.0\%,
84.4\%, and 86.7\% for $K=30$, 45, and 60, respectively. Increasing the
context beyond 60 provides little additional improvement, with terminal means
of 0.813 and 0.862~km for $K=75$ and 90. Meanwhile, batched policy-call latency
increases from 10.98~ms at $K=15$ to 14.35~ms at $K=90$. These results support
a performance-computation tradeoff with diminishing returns near
$K=45$, rather than unlimited gains from larger context. Retaining more
than the first few ranked objects leaves the policy scope to select a
lower-ranked candidate when geometry or future value favours it. 

\subsubsection{Representation and Attention Diagnostics}
Unlike the distinct Phase-I inclination branch, Phase-II scenarios in
Fig.~\ref{fig:representation-diagnostics}(a) overlap along catalogue-quality
gradients, with recurrent states more continuous than encoder summaries.
Projection-intrusion rates of 6.1-27.6\% warrant caution when interpreting
local separations. In Fig.~\ref{fig:representation-diagnostics}(b), attention
is more strongly associated with slew than in Phase I. This is consistent
with greater emphasis on relative geometry and economical retargeting
among the more numerous opportunities in a denser perceptive field. Teammate tokens receive 14.3\% of attention
from the acting sensor's query and 15.4\% from RSO queries, near the 15.3\%
uniform-attention reference. Attention alone, however, does not establish
the performance benefit of teammate information.

\subsection{Phase III: Heterogeneous Cooperation under Overlapping Coverage}
\label{sec:results-phase3}
\subsubsection{Complementary Allocation and Coverage}

\begin{figure}[!h]
\centering
\includegraphics[width=\columnwidth]{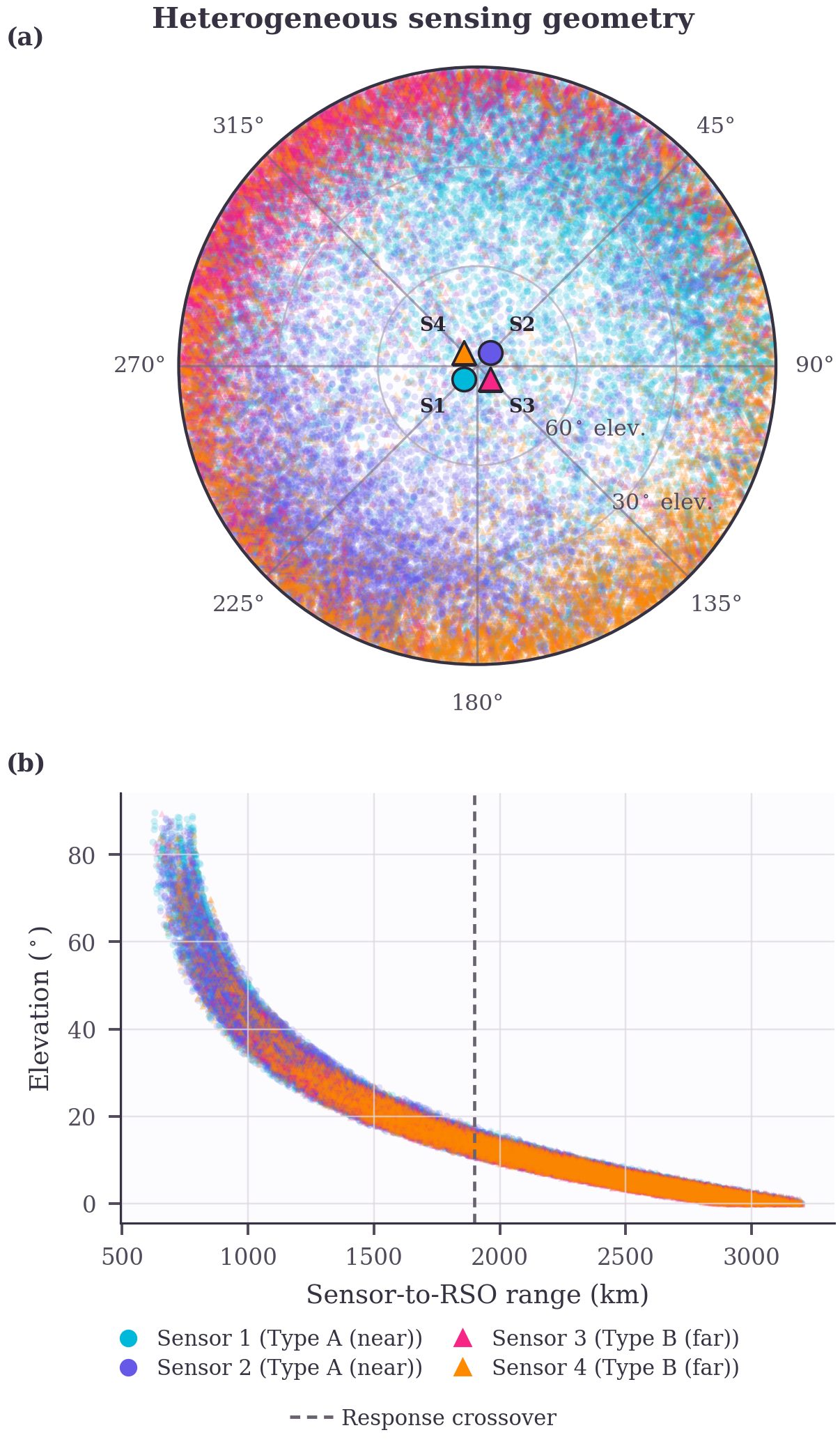}
\caption{Phase-III heterogeneous observation geometry. The figure shows a
50,000-event random subsample of 60,484 successful observations collected over
six episodes. \textbf{(a)} Polar map of the four superimposed sensor-local sky
domes. Azimuth is angular and elevation increases toward the centre. Colour
identifies the individual sensor, while circles and triangles identify Type-A
and Type-B sensors, respectively. The labelled S1-S4 symbols show the angular
offsets of the ground sites from the network centre; observation coordinates
remain expressed in each observing sensor's local frame. \textbf{(b)}
Elevation versus sensor-to-RSO range for the same observations. The dashed
line marks the 1900-km crossover between the two sensor response models.}
\label{fig:phase3-observation-geometry}
\end{figure}

Within the four sensors' highly overlapping perceptive fields,
Fig.~\ref{fig:phase3-observation-geometry} shows allocation patterns consistent
with their complementary capabilities. Mean successful-observation ranges are
1550.84 and 1636.60~km for the two Type-A sensors and 2312.02 and 2398.87~km for
the two Type-B sensors. This separation is consistent with the complementary
response functions: Type A is favoured at shorter range and Type B at longer
range, with equal quality at 1900~km. The policy is not supplied with the
analytic response laws, but it observes geometry, sensor type, angular
proximity, and tasking history, allowing it to learn capability-dependent
allocation.

Across 12 evaluation episodes, the mean pairwise nonduplication rate is
$98.27\pm0.15\%$, and all 240 objects are observed at least once in every
episode. The episode-averaged effective coverage defined in
Eq.~\eqref{eq:effective-coverage} is $85.76\pm0.95\%$. Thus, the shared policy
distributes sensing effort broadly across the complete catalogue while avoiding
unnecessary simultaneous assignments.

\subsubsection{Modality Dependence}

To test whether sensor type functionally affects the learned policy, the
modality embeddings are evaluated under three inference conditions: the
correct labels, a neutral label formed from the mean learned embedding, and
exchanged Type-A and Type-B labels. Physical response functions, policy
weights, and paired environment seeds remain unchanged.

The mean endpoint uncertainties are
0.463735, 0.526422, and 0.815580~km for the correct, neutral, and exchanged
conditions. Neutralizing the labels degrades the mean by 13.518\%, while
exchanging them produces a 75.872\% degradation. The intervention therefore
shows that modality information materially conditions the learned allocation.
Together with the range separation, low duplication, and broad coverage, this
supports complementary capability-aware tasking within the heterogeneous
sensing regime.

%% file: Sections/conclusion.tex
This work introduced VISTA to address a central representational limitation of
deep reinforcement learning sensor tasking: fixed observation and action
interfaces cannot naturally accommodate changing target identities, catalogue
sizes, or sensing networks. VISTA instead retrieves a bounded, physics- and
mission-informed candidate set and represents sensors and candidate RSOs as
typed entities. Attention models their relationships, recurrent memory carries
temporal context, and pointer decoding selects through the current candidate
representations. This design
preserves the correspondence between physical objects and actions while
allowing one shared policy to operate across variable catalogue and network
dimensions.

The three experimental campaigns provide complementary evidence for this
formulation. In the fixed 30-RSO benchmark, where variable-cardinality and
multi-agent capabilities are not required, VISTA reaches the recovery
threshold 31.2\% faster than the flat recurrent baseline. In the LEO-to-LEO
scenario with 12 orbital sensors and 2000 RSOs, defined by high-dimensionality and complex relative movement, its five-hour mean uncertainty
is 97.5\% below EIG and 99.3\% below the recurrent baseline. Frozen-policy
experiments extending to 20,000 RSOs reveal
an approximately linear relation between catalogue size and the smallest
successful tested network for a given temporal horizon. In the heterogeneous campaign, the agents allocate
observations according to sensor capability while maintaining broad catalogue
coverage. 

These experiments were designed to compare optimization methods and neural
representations rather than reproduce operational catalogue accuracy. The
models were deliberately simplified to enable large-scale scenarios, repeated
evaluations, broad comparisons, and statistically meaningful analysis.
Absolute uncertainty values and empirical capacity relations are therefore
conditional on these abstractions. Applying the same models and interfaces to
every method nevertheless isolates relative scheduling and representational
capabilities. The resulting gains, particularly over the
fixed-dimensional recurrent architecture traditionally used in SSA-oriented
DRL, constitute the central evidence of this study.

VISTA consequently broadens the architectural basis available for learned SSA
tasking by adapting entity attention and pointer selection from other domains
while retaining a temporal core. It should be regarded as an initial scalable
solution and a reproducible basis for further development. Future work should
integrate it into higher-fidelity SSA chains with richer orbital dynamics,
sensing and estimation, imperfect information exchange, and broader geometric
variation. Controlled ablations can then clarify which components drive the
gains and whether they persist as operational fidelity increases.

\newpage

%% file: Profiles/biography.tex
\begin{IEEEbiography}
    [{\includegraphics[width=1in,height=1.25in,keepaspectratio]{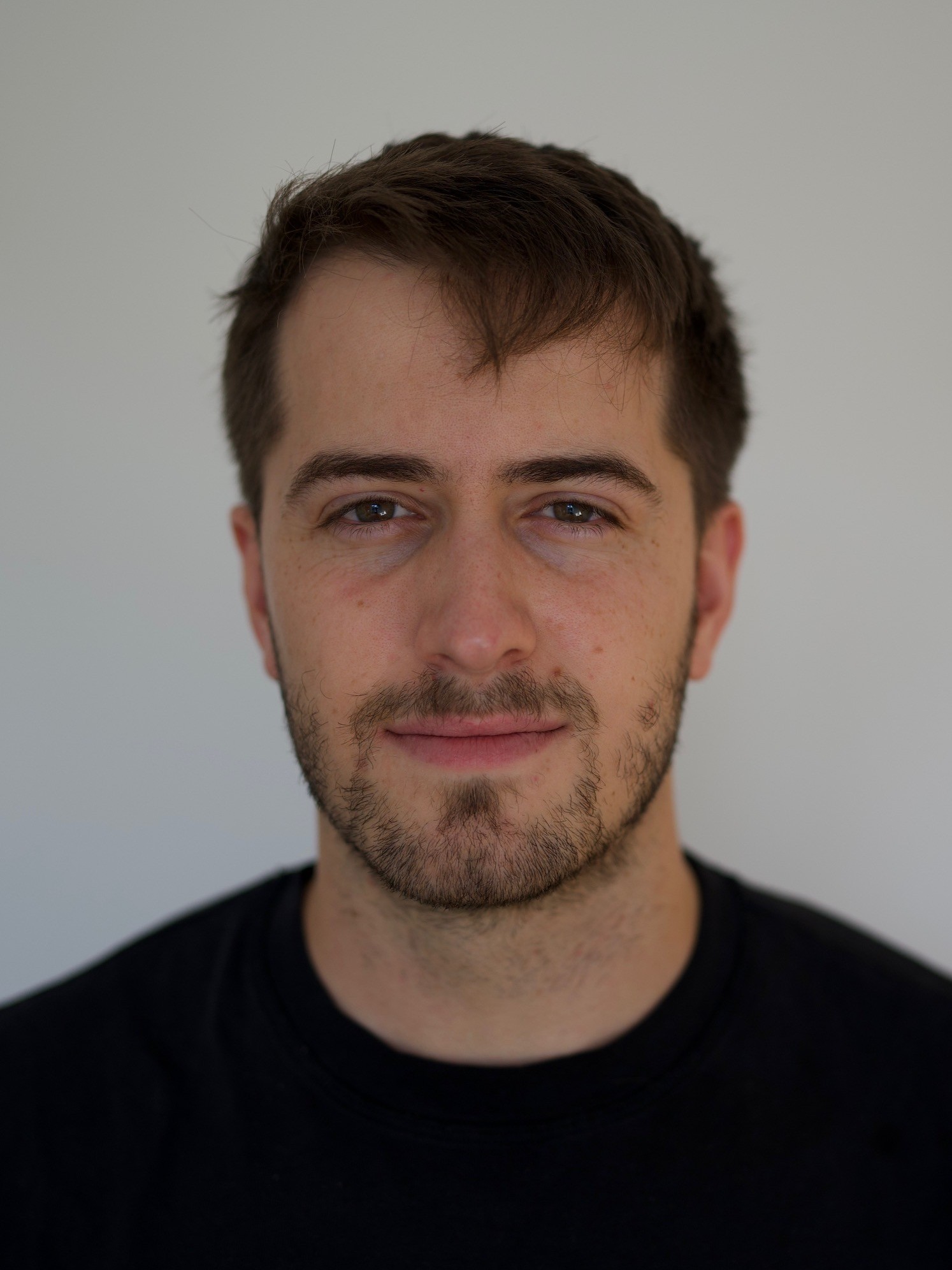}}]
    {Miguel Leiva-Vélez}
    received the B.S. degree in aerospace engineering from Universidad Politécnica de Madrid, Madrid, Spain, in 2022, and the M.S. degree in space engineering from Universidad Carlos III de Madrid, Madrid, Spain, in 2024. He is currently pursuing the Ph.D. degree in aerospace engineering at Universidad Politécnica de Madrid.

    He previously researched deep-learning-based optical navigation for proximity operations around noncooperative spacecraft. He is currently a Senior Researcher in deep reinforcement learning at Indra Sistemas, Madrid, Spain. His research interests include scalable optimization and intelligent decision-making for aerospace systems, particularly space situational awareness, planetary landing, and guidance, navigation, and control.
\end{IEEEbiography}

\begin{IEEEbiography}
    [{\includegraphics[width=1in,height=1.25in,keepaspectratio]{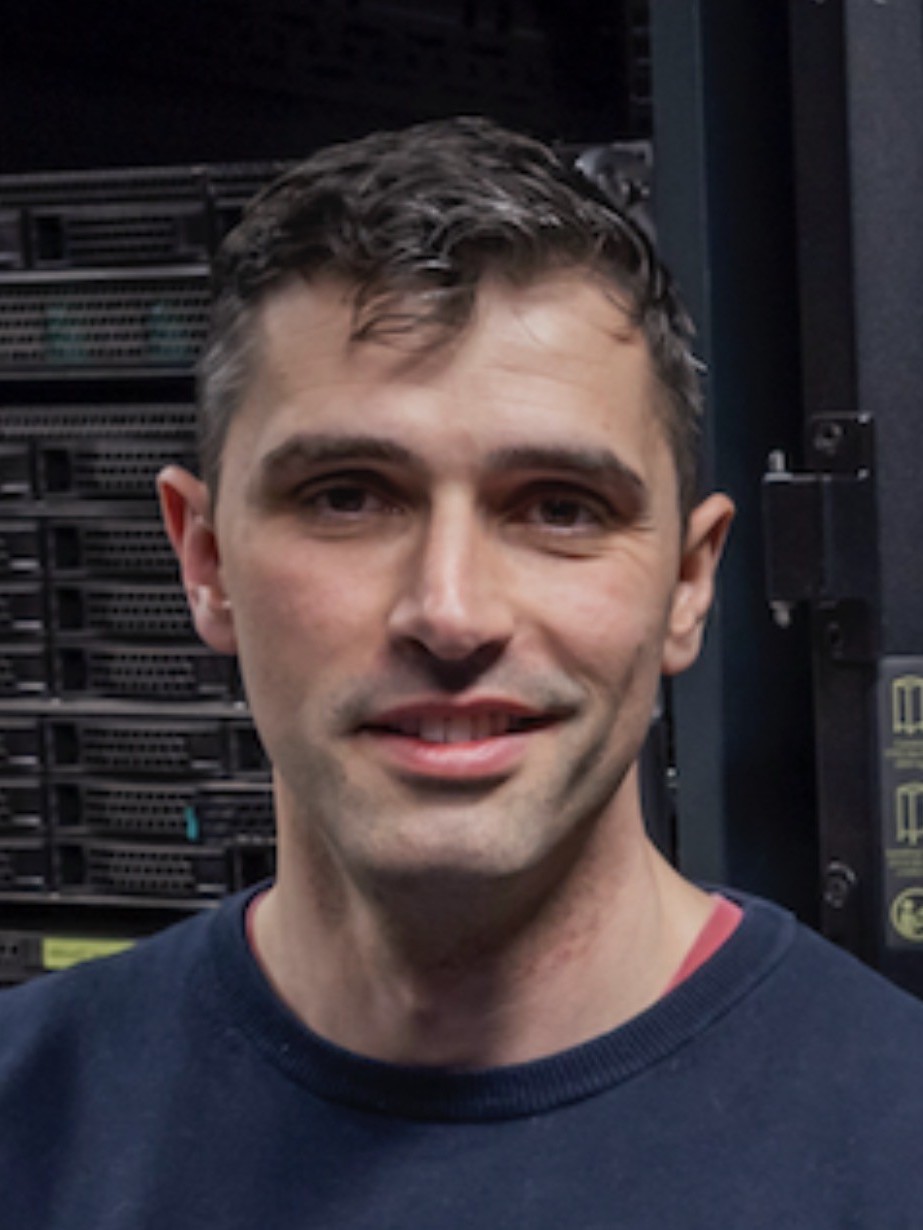}}]
    {Adalberto Claudio Quiros}
    received in 2013 the M.S. degree in electrical engineering from the Illinois Institute of Technology, Chicago, IL, USA, and the M.S. and B.S. degrees in telecommunications engineering from the Universidad Politécnica de Madrid, Madrid, Spain. He received the Ph.D. degree in machine learning and computer science from the University of Glasgow, Glasgow, Scotland, U.K., in 2022.

    He is a Staff Machine Learning Scientist at Indra Sistemas, Madrid, Spain. Previously, he held Lecturer and Research Associate positions at the University of Glasgow and was a System-on-Chip Design Engineer at Altera Corporation and Intel Corporation. His research interests include multimodal perception, representation learning, interpretable artificial intelligence, and edge-efficient machine-learning systems.
\end{IEEEbiography}

\begin{IEEEbiography}
    [{\includegraphics[width=1in,height=1.25in,keepaspectratio]{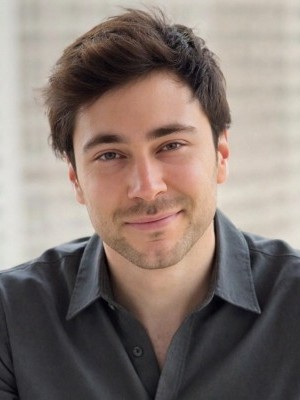}}]
    {Nicolas Gaston Rozado}
    received the B.S. degree in psychology from Universidad Complutense de Madrid, Madrid, Spain, in 2016, and the M.S. degree in theoretical neuroscience from Universitat Pompeu Fabra, Barcelona, Spain, in 2017.

    He is Head of Artificial Intelligence for Defense and Safety-Critical Systems at Indra Sistemas, Madrid, Spain. His previous research focused on brain modelling and complex dynamics. His current research interests include deep reinforcement learning and autonomous decision-making.
\end{IEEEbiography}

\begin{IEEEbiography}
    [{\includegraphics[width=1in,height=1.25in,keepaspectratio]{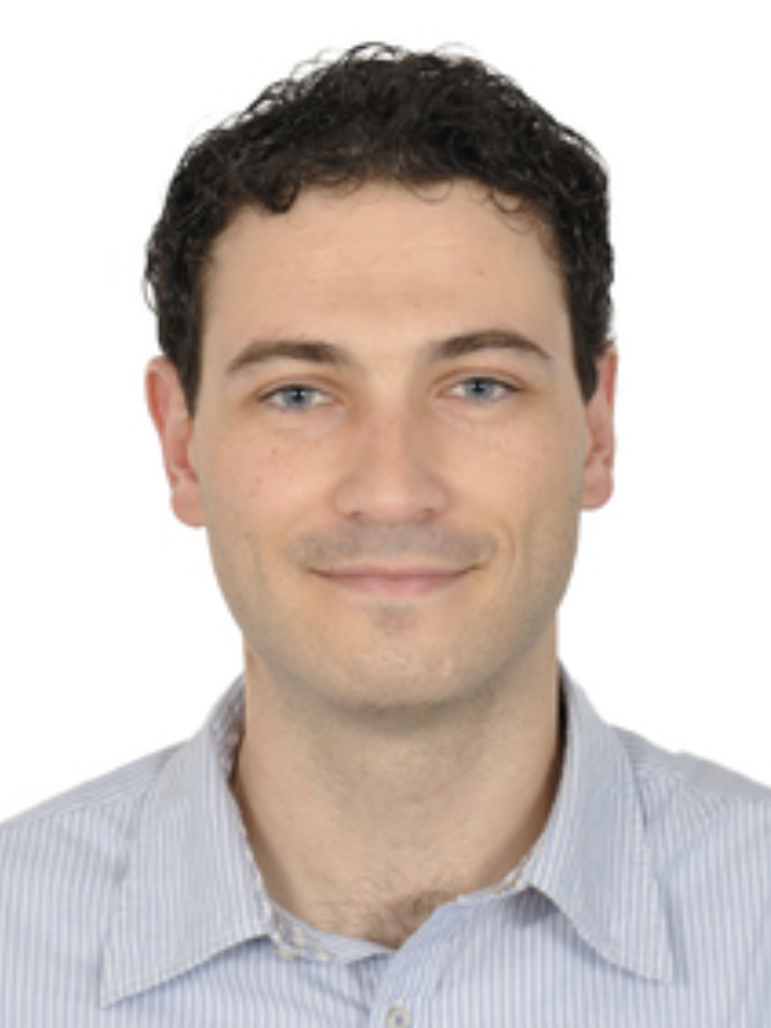}}]
    {Hodei Urrutxua}
    received the M.S. and Ph.D. degrees in aerospace engineering from Universidad Politécnica de Madrid, Madrid, Spain, in 2011 and 2015, respectively.

    He is an Associate Professor of Aerospace Engineering at Universidad Rey Juan Carlos, Madrid, Spain. His research interests include high-performance numerical techniques for orbit and attitude propagation; space situational awareness, including asteroid-capture dynamics, slow-deflection techniques, and active space-debris removal; and spacecraft guidance, navigation, and control, with an emphasis on AI-enhanced methods.
\end{IEEEbiography}

\begin{IEEEbiography}
    [{\includegraphics[width=1in,height=1.25in,keepaspectratio]{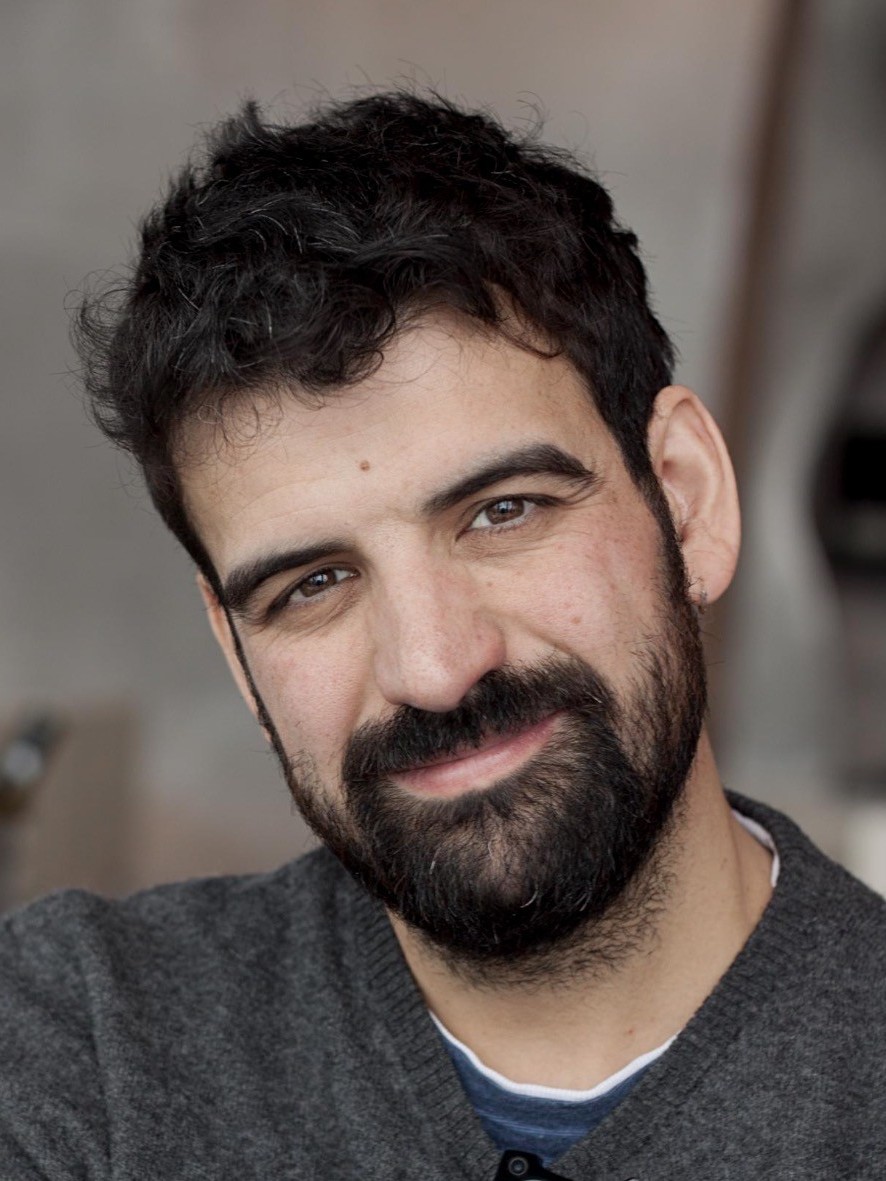}}]
    {Víctor Rodríguez-Fernández}
    received the B.S. degree in mathematics and computer science in 2013, the M.S. degree in information technology in 2015, and the Ph.D. degree in computer science in 2019, all from Universidad Autónoma de Madrid, Madrid, Spain.

    He is an Associate Professor of Computer Science at Universidad Politécnica de Madrid, Madrid, Spain, and a Research Affiliate with the Astrodynamics, Space Robotics and Controls Laboratory, MIT AeroAstro, Cambridge, MA, USA. He collaborates with NASA's Jet Propulsion Laboratory and the European Space Agency. His research interests include time-series analysis, large language models, and AI agents.
\end{IEEEbiography}